\documentclass[twoside]{article}

\usepackage{caption}

\usepackage{algorithmic}
\usepackage{amsmath}
\usepackage{amssymb}
\usepackage{mathtools}
\usepackage{amsthm}
\usepackage{amsmath}
\usepackage{dsfont}
\usepackage{graphicx}
\usepackage{caption}
\usepackage{multirow}
\usepackage{pgfplots}
\usepackage{algorithm}
\usepackage{algorithmic}
\usepackage{booktabs} 
\usepackage[utf8]{inputenc} 
\usepackage[T1]{fontenc}    
\usepackage{url}            
\usepackage{amsfonts}      
\usepackage{nicefrac}       
\usepackage{microtype}     
\usepackage{xcolor}  

\newcommand{\secondbest}[1]{\underline{#1}}
\usepackage{tikz}
\usepackage{calc}

\newcommand{\wavyunderline}[1]{%
  \tikz[baseline=(X.base)]{
    \node[anchor=base, inner sep=0pt] (X) {#1};
    \draw[decorate, decoration={snake, amplitude=0.5pt, segment length=3pt}]
      (X.south west) -- (X.south east);
  }%
}

\newcommand{\thirdbest}[1]{\wavyunderline{#1}}

\newcommand{\dd}{\operatorname{d}\!}
\newcommand{\dataset}{{\mathcal D}}

\def\T{^{\top}} 

\usepackage{caption}
\usepackage[capitalize,noabbrev]{cleveref}
\usepackage{enumitem}

\usepackage[preprint]{aistats2026}
\usepackage[round]{natbib}

\begin{document}

%

%

\twocolumn[

\aistatstitle{From Overfitting to Reliability: Introducing the
Hierarchical Approximate Bayesian Neural Network}

\aistatsauthor{Hayk Amirkhanian \And Marco F. Huber}

\aistatsaddress{ \textit{Fraunhofer IPA} \\
\textit{University of Stuttgart}\\
Stuttgart, Germany \\
hayk.amirkhanian.namagerdi@ipa.fraunhofer.de \And  \textit{Fraunhofer IPA} \\
\textit{University of Stuttgart}\\
Stuttgart, Germany \\
marco.huber@ieee.org } ]

\begin{abstract}
In recent years, neural networks have revolutionized various domains, yet challenges such as hyperparameter tuning and overfitting remain significant hurdles. 
Bayesian neural networks offer a framework to address these challenges by incorporating uncertainty directly into the model, yielding more reliable predictions, particularly for out-of-distribution data.
This paper presents \emph{Hierarchical Approximate Bayesian Neural Network (HABNN)}, a novel approach that uses a Gaussian-inverse-Wishart distribution as a hyperprior of the network's weights to increase both the robustness and performance of the model.
We provide analytical representations for the predictive distribution and weight posterior, which amount to the calculation of the parameters of Student's $t$-distributions in closed form with linear complexity with respect to the number of weights. 
Our method demonstrates robust performance, effectively addressing issues of overfitting and providing reliable uncertainty estimates, particularly for out-of-distribution tasks. 
Experimental results indicate that HABNN not only matches but often outperforms state-of-the-art models, suggesting a promising direction for future applications in safety-critical environments.
\end{abstract}

\section{Introduction}
Neural networks—backed by successes in medical diagnostics \citep{med_diag_AI}, autonomous control \citep{aut_contrl_AI}, mass production \citep{int_mass_prod_AI}, and vision tasks \citep{collobert2008unified}—are typically trained by backpropagating scalar weights \citep{backprop_alg}. 

Yet they still suffer from overfitting \citep{neural_netw_overfit}, extensive hyperparameter tuning, and unreliable uncertainty estimates \citep{PBP,BNN_unreliable_1,BNN_unreliable_2}, which is critical in high-risk domains.

Bayesian neural networks (BNNs) promise principled uncertainty modeling \citep{Bayes_NN1,Bayes_NN2} but are hampered by intractable posteriors. 
Approximations like PBP \citep{PBP}, TAGI \citep{TAGI}, SVI \citep{SVI}, and KBNN \citep{KBNN} usually rely on Gaussian assumptions that misestimate uncertainty—especially out-of-distribution (OOD)—and sometimes incur quadratic costs when modeling weight covariances.
These unreliable uncertainty estimates—especially for out-of-distribution (OOD) inputs—have been shown to worsen when standard data-normalization steps (e.g., zero-mean/unit-variance scaling or batch-normalization) strip away input-dependent variance cues, leading to overconfident predictions under covariate shift \citep{normalization_unreliable}.

In order to account for those issues, we propose the \emph{Hierarchical Approximate Bayesian Neural Network (HABNN)}, a gradient-free inference method providing analytic posterior updates. 
Our forward pass yields predictive distributions; the backward pass updates weight distributions; and we support online learning without batch updates. 

We demonstrate HABNN’s robustness on hyperparameter initialization, reliable uncertainty estimations on OOD data, and state-of-the-art performance on UCI regression benchmarks and the Industrial Benchmark \citep{SIB}.

\newpage

Our contributions include:~
\begin{itemize}
    \item Introducing HABNN, a gradient-free, online Bayesian inference method for analytical posterior estimation.
    \item Deriving analytical expressions for both the forward and backward passes.
    \item Demonstrating HABNN’s effectiveness on UCI regression datasets and the Industrial Benchmark.
\end{itemize}

\section{Related Work}
\label{sec:relatedwork}
BNN inference spans several approaches, each trading off accuracy, complexity, and uncertainty fidelity.

Markov Chain Monte Carlo methods—including Metropolis–Hastings, Gibbs sampling \citep{gibbs}, Hamiltonian Monte Carlo \citep{HMC,hybridMC}, and the No-U-Turn Sampler \citep{nuts}—are the gold standard (\citep{MCMC}) but suffer from high cost and lack closed-form posteriors.

Variational Inference (VI) \citep{VI} approximates the true posterior with a simpler surrogate distribution (usually Gaussian), minimizing the evidence lower bound (ELBO) via gradient descent \citep{SGD}; its stochastic variants—SVI \citep{SVI} and Monte Carlo dropout \citep{Dropout}—scale to large datasets, with recent extensions \citep{subspace_VI,function_VI}. 

Expectation Propagation \citep{EP}, typified by Probabilistic Backpropagation \citep{PBP}, approximates the predictive distribution to enable analytic forward-pass moment propagation, though its backward-pass updates differ from VI in the sense that PBP minimizes the reverse Kullback-Leibler divergence instead of the forward divergence as VI does. 

The Laplace approximation \citep{orig_laplace} fits a Gaussian distribution around the maximum a posteriori (MAP) estimate via a Taylor-series expansion (e.g., \citep{backpack_laplace,riemannian_Laplace}). 

Since the 1990s, Bayesian filtering techniques like the EKF \citep{EKF} or UKF \citep{UKF}, which are based on the famous Kalman filter \citep{Kalman_Filter}, are examined for training BNNs \citep{watanabe1990, layer_indp}. This has recently led to TAGI \citep{TAGI} and KBNN \citep{KBNN}, which assume Gaussian weight distributions resulting in closed-form update equations for the mean and covariance. TAGI and KBNN differ in covariance assumptions and update rules. 

Several recent works have underscored the advantages of heavy‐tailed priors in BNNs, with \cite{Stud_t_BNN_1} showing that infinite‐width BNN posteriors converge to Student’s \(t\) processes and \cite{Stud_t_BNN_2} demonstrating practical benefits of Student’s \(t\) priors over Gaussians. 
More recently, \cite{Stud_t_BNN_3} introduced Deep Evidential Regression, training a deterministic network to output Normal–Inverse–Gamma parameters that induce a Student’s \(t\) predictive distribution and naturally decompose aleatoric and epistemic uncertainty.

In this work, we generalize TAGI and KBNN, which both assume Gaussian weight distributions, by modeling weights with Student’s $t$-distributions, endowing heavier tails for robustness and reliable OOD uncertainty. As data accumulates, these $t$-distributions naturally converge to Gaussians, reducing initialization sensitivity while matching or exceeding current state-of-the art BNN performance.

\section{Problem Formulation}
\label{sec: Problem Formulation}
Assume a dataset $\dataset = \{(x_j, y_j)\}_{j=1}^{N}$ of $N$ independent and identically distributed (i.i.d) samples, with inputs $x_i \in \mathds{R}^d$ and outputs $y_i \in \mathds{R}$.
Further, assume that $y_i$ is given by $y_i = f_{W}(x_i) + \epsilon$, where $f_W$ is a multi-layer (fully-connected) BNN with rectified linear unit (ReLU)\footnote{The proposed approach can be straightforwardly extended to general piece-wise linear activations, e.g., leaky ReLU.} \citep{ReLU} activation functions in all layers except for the output layer, where no activation is applied.
The weights $W$ are the parameters of the network.
The term $\epsilon \sim \mathbb{N}(0, \sigma_{\epsilon}^{2})$ will be employed as it accounts for the aleatoric uncertainty of the data, where $\mathbb{N}(\mu, \sigma^2)$ is a Gaussian distribution with mean $\mu$ and variance $\sigma^2$. 
Remember that in a BNN, the weights $W$ are random variables unlike the deterministic neural network setting in which the weights are deterministic scalar values. 
The BNN consists of $l=1,\ldots,L$ layers, each of which is given by
\begin{align}
    z^{l+1} = f(a^l)~, \text{ and }  
    a^l = \frac{W^l \cdot z^l + b^l}{\sqrt{M_{l+1}}} \label{eq:layer_def}~,
\end{align}
with $b^l \in \mathds R$ and $W^l \in \mathds R^{M_{l+1} \times M_l}$, where $M_{l+1}$ is the number of neurons of layer $l$ and $W = \{W^l\}_{l=1}^L$. Furthermore, $z^l \in \mathds{R}^{M_l}$ with $z^1 = x$ for the input layer and $z^{L+1} = y$ for the output layer. The ReLU activation function $f(\cdot)$ is applied element-wise.
Please note that in order to ease the notation, the bias $b^l$ will be included into the weight matrix $W^l$ in the following.

\begin{table*}[t]
\caption{Average OOD NLL (RMSE) relative error (in \%) when normalizing the training data.}
\centering
\small
\sc
\setlength{\tabcolsep}{3pt}
\begin{tabular}{lccccccc}
\toprule
Dataset & PBP & Dropout & SVI & KBNN & TAGI & MCMC & Laplace \\
\midrule
Concrete & 34 (327) & 55 (33) & 472 (971) & 93 (949) & 679 (164) & 172 (5131) & 86 (234) \\
Energy & 8 (267) & 55 (35) & 283 (1991) & 67 (1235) & 65 (523) & 475 (4969) & 83 (270)\\
Wine & 59 (3635) & $>10^6$ (15886) & 347 (8914) & 62 (27297) & 130 (7413) & 390 (122) & 31 (275) \\
Naval & $>10^6$ ($>10^6$) & $>10^6$ (41967) & 468 (34445) & 336 ($>10^6$) & 12516 ($>10^6$) & 42330 (651) & 184 (400)\\
Yacht & 62 (200) & 7 (4) & 2709 (7589) & 78 (1155) & 75 (637) & 13758 (15803) & 93 (72)\\
Kin8nm & 3114 (63) & 31611 (156) & 367 (261) & 808 (248) & 2939 (117) & 646 (82) & 928 (218)\\
\bottomrule
\bottomrule
Average & $>10^6$ ($>10^6$) & $>10^6$ (9680) & 774 (9028) & 241 (23402) & 2734 ($>10^6$) & 9628 (4460) & 234 (245)
\end{tabular}
\label{tab: OOD Rel Error}
\end{table*}

\newpage

The difficulty of estimating the parameter $W$ of a BNN lies in solving both the posterior
\begin{align}
    p(W \, | \, \dataset) = \frac{p(\dataset \, | \, W) \, p(W)}{p(\dataset)} \label{eq:bayes_posterior}
\end{align}
and the predictive distribution 
\begin{align}
    p(y_* \, | \, x_*, \dataset) = \int_{\mathds R^{M_{l+1}\times M_l}} p(y_* \, | \, x_*, W) \, p(W \, | \, \dataset) \, \dd W~, \label{eq:predictive_distr}
\end{align}
where $y_*, x_*$ describe a new, so far unseen output and input, respectively.

Many methods approximate these distributions (cf. Sec.~\ref{sec:relatedwork}), but most rely on normalizing the training data\footnote{E.g.\ min–max, mean, or z‐score normalization (standardization).} to perform well.
However, normalization can remove the very cues Bayesian models need for reliable epistemic uncertainty \citep{normalization_unreliable}.

To assess BNN reliability under OOD data, we compute the average relative error over three OOD scenarios for both NLL and RMSE:
\begin{align}
  \Delta_{\mathrm{NLL}} &= \frac{1}{3}\sum_{i=1}^3 \frac{|\mathrm{NLL}_i - \mathrm{NLL}_{\mathrm{in}}|}{|\mathrm{NLL}_{\mathrm{in}}|}\times100\%~, \label{Eq: nll_ood}\\
  \Delta_{\mathrm{RMSE}} &= \frac{1}{3}\sum_{i=1}^3 \frac{|\mathrm{RMSE}_i - \mathrm{RMSE}_{\mathrm{in}}|}{|\mathrm{RMSE}_{\mathrm{in}}|}\times100\%~, \label{Eq: rmes_ood}
\end{align}
where the index $i$ indicates the OOD test sets, which are obtained by scaling the inputs by 0.1 or 2 as well as by adding $3\times$ their standard deviation. 

As Table~\ref{tab: OOD Rel Error} shows, none of the models yield reliable uncertainty estimates (or actual predictions) when faced with OOD data—critical, since BNNs are often deployed in high‐risk settings where trust in predictions is paramount.

\vspace{2cm}
In the next section, we derive the \emph{Hierarchical Approximate Bayesian Neural Network (HABNN)}, which matches state‐of‐the‐art in‐distribution performance and offers more robust and reliable OOD performance, analytical forward \eqref{eq:predictive_distr} and backward \eqref{eq:bayes_posterior} updates, supports online learning, and is computationally efficient.

\section{Hierarchical Approximate Bayesian Neural Network}
Since the training data is assumed to be i.i.d., each sample can be processed individually. 
We will assume a weak Markov property of (Bayesian) neural networks as suggested by \citet{layer_indp} and done by \citet{KBNN} and \citet{TAGI}.
Therefore the input of any given layer is the output of the previous layer and vice versa.
Importantly, the sizes or outputs of earlier layers (other than the immediately preceding layer) do not influence the output of the current layer. 
This significantly eases the calculations of the weights as one can now train not only every layer on its own but also to simply focus on the distribution of the pre-activation $a$ and the post-activation $z$ in \eqref{eq:layer_def}. 
Thus, training is performed recursively, where the posterior of the previous training pass becomes the prior for the current iteration.
The remainder of this section is organized as follows: 
Sec.~\ref{sec:habnn_weight} introduces the weight distribution in our model. 
Sec.~\ref{sec: Foward Pass} presents the forward pass for the proposed network.
Finally, Sec.~\ref{sec: Backward Pass} derives the backward pass, applying moment matching to approximate the intractable posterior.

\subsection{Weight Distribution}
\label{sec:habnn_weight}
Many BNNs normalize their inputs (e.g., min–max or z-score), but this suppresses the outliers that signal epistemic uncertainty \citep{normalization_unreliable}. Because most methods assume Gaussian weight priors \citep{SVI,KBNN,VI}, they remain vulnerable to outliers and initialization, which limits the exploration of the weight space.

\newpage

Instead, we use a hierarchical Gaussian–Inverse‐Gamma prior that marginalizes to a heavy‐tailed Student’s $t$ distribution \citep{ConjugatePrior}, and extend this to layerwise vectors via a Gaussian–Inverse‐Wishart hyperprior for multivariate $t$’s. 
Since the $t$ distribution does not belong to the exponential family, we apply assumed‐density filtering—approximating both layer outputs and weight posteriors as (multivariate) Student’s $t$—and adopt a mean‐field simplification, following \citet{TAGI}’s finding that posterior covariances effectively diagonalize.
For a more detailed discussion of the weight distribution, see Appendix~\ref{sec:Appendix Weight Distribution}.

\subsection{Forward Pass}
\label{sec: Foward Pass}
Note that due to treating the layers sequentially, we will omit all layer indices to increase readability.
In theory a rank-reducing linear transformation of a multivariate Student's t-distribution would result in a Student's t-distribution again \citep{ConjugatePrior}, however after the first layer the multiplicative non-linearity in Equation \eqref{eq:layer_def} distorts this.
In order to overcome this issue, it will be assumed that $p(a | D)$, i.e., the distribution of the pre-activation $a$, will always be Student's t-distributed for any given layer.
In the following, a moment matching approach will be employed to estimate the parameters of the distribution.
Luckily, both the mean and scale matrix have been derived by \citet{KBNN} for the Gaussian case already. These derivations also hold for Student's t-distributions and therefore we will simply state the final formulae. 

The elements of the mean vector are given by
\begin{align}
    (\mu_a)_i = \frac{(\mu_W\T)_i \, \mu_z}{\sqrt{M_{l+1}}} \quad \forall i \in M_{l+1}~, \label{eq:mean_a}
\end{align}
where $\mu_z$ is the mean of the post activation $z$ of the previous layer, defined in Equation~\eqref{eq:layer_def}.
Due to the mean-field assumption, both the scale and covariance matrix are diagonal.

The elements of the diagonal of the covariance matrix are then given by
\begin{align}
    (\sigma_a^2)_i = \frac{(\mu_W\T)_i C_z (\mu_W)_i\hspace{-.5mm} +\hspace{-.5mm} \mu_z\T C_{W_j} \mu_z\hspace{-.5mm} +\hspace{-.5mm} \mathrm{Tr}(C_{W_i} C_z)}{M_{l+1}}    \label{eq:cov_a}
\end{align}
with Tr($\cdot$) denoting the trace of a matrix, and $C_W$ and $C_z$ representing the covariance matrices of the weights $W$ and the post-activation values $z$, respectively.

For a multivariate Student's t-distribution the covariance matrix is not equal to the scale matrix but instead only defined if the degrees of freedom are greater than two.

In that case the covariance matrix can be calculated according to 
\begin{align}
    C = \frac{\nu}{\nu-2} \, \Sigma~, \label{eq:scale_matr}
\end{align}
with $\Sigma$ being the scale matrix, $\nu$ the degrees of freedom, and $C$ the covariance matrix.
This implies that, in our case, the diagonal elements $\tau_a^2$ of the scale matrix of $a$ can be computed as
\begin{align*}
    (\tau_a^2)_i = \frac{v_a - 2}{v_a} (\sigma_a^2)_i~.
\end{align*}

In the first layer, the degrees of freedom remain unaffected by the linear transformation, so it was decided not to alter them in subsequent layers either.

This concludes the pre-activation calculations. 
The next step is the element-wise application of the ReLU function to the pre-activation $a$ as defined in \eqref{eq:layer_def}.  
Assuming a diagonal covariance structure, the mean and variance of each post-activation $z = f(a)$ can be computed independently and later vectorized. 
Both the mean and scale are calculated analytically without any approximations.  
The detailed derivations can be found in Appendix~\ref{sec:Appendix Forward Pass}; here, we only state the final results.  
For clarity, we demonstrate the results for a single neuron $z_i = f(a_i)$, $i=1,\ldots, M^{l+1}$, omitting the index for readability.  
We also assume the parameters of $a$ are given by $\mu, \tau^2, \nu$.

The mean value and variance of $f(a)$ are then given by
\begin{align}
\mathbb{E}[\, \mathrm{f}(a)] =: \mu_z &= \frac{\Gamma(\frac{\nu+1}{2})}{\Gamma(\frac{\nu}{2}) \sqrt{\pi \nu \tau^2}} \, \frac{\nu \tau^2}{\nu-1} \left(1 + \frac{\mu^2}{\nu \tau^2}\right)^{\frac{1-\nu}{2}} \nonumber \\ 
&+ \mu \, (1 - \text{F}(\, 0\, | \, \mu,\tau^2, \nu))~, \label{eq:exp_val_f(a)} \\
\text{Var}[\, \mathrm{f}(a)] &=: \sigma_z^2 = \frac{\nu \tau^2}{2 (\nu-2)}
+ \mu^2 \, \text{F}(\, \mu \, | \, 0,\tau^2, \nu) \nonumber \\
&+ \mathrm i \hspace{0.15cm} \frac{\nu \tau^2}{2 \sqrt{\pi}} \hspace{0.15cm} \frac{\Gamma(\frac{\nu+1}{2})}{\Gamma(\frac{\nu}{2})}  \hspace{0.15cm} \mathbf B_{\frac{\mu^2}{\nu \tau^2 + \mu^2}}\left(\frac{3}{2}, \frac{\nu-2}{2}\right) \nonumber \\[0.2 cm] \nonumber \\
&+ 2 \mu \, (\mu_z - \mu \ (1 - \text{F}(\, 0 \, | \, \mu,\tau^2, \nu))) - \mu_z^2~, \label{eq:cov_f(a)}
\end{align}
respectively, where $\Gamma$ is the Gamma function, F denotes the cumulative distribution function of a Student's t-distributed random variable, and $\mathbf B$ is the Beta function, defined in the appendix in~\eqref{eq:beta_fct}.

\newpage

\subsection{Backward Pass}
\label{sec: Backward Pass}
This section shifts focus from predicting $y$ given an input $x$ to updating the weights $W$ using training data $\dataset$. 
Under the assumption that training samples are i.i.d. and making use of the Markov property of (Bayesian) neural networks, weight updates can be performed in sequence. 
Consequently, samples are processed one at a time, eliminating the need for batch processing typically seen in neural network training.
In this context, the posterior from the previous sample is used as the prior for the next one.

Updating the weights of a layer is performed in two steps: first, $a_{l}$ gets updated using $z_{l+1}$ and then,  $\left(W_l\T, z_l\T\right)\T$  gets updated using $a_{l}$.
Since we derive the update equations for both steps similarly, we denote both involved variables of each step as $X_1$ and $X_2$. 
We assume that they are jointly multivariate Student's t-distributed and we calculate the posterior $p(X_1 | \dataset)$ by marginalizing over $X_2$, yielding
\begin{align}
    p(X_1 | \dataset) &= \int_{\Omega_{X_2}} p(X_1 \, | \, X_2) \, p(X_2 \, | \, \dataset) \, \dd X_2~.  \label{eq:posterior}
\end{align}

Unlike the Gaussian case, where the posterior \eqref{eq:posterior} can be computed analytically \citep{KBNN}, this is not possible here because the product of two multivariate Student's~$t$ probability density functions (PDFs) does not yield another multivariate Student's~$t$ PDF.
To address this issue, we approximate the true posterior in Equation \eqref{eq:posterior} by using a surrogate multivariate Student's $t$-distribution, matching its moments to those of the true posterior. 
Similar to the forward pass, the calculations can now be done analytically without any approximations. 
We will again only state the final update formulae for the mean, scale matrix as well as the degrees of freedom. 
Interested readers are referred to Appendix~\ref{sec: appendix Backward Pass} for the detailed derivations.
The update formulae for the mean, scale matrix, and degrees of freedom are given by
\begin{align}
    \mathbb{E}[X_1 \, | \, \dataset] 
&= \mu_1 + \Sigma_{12} \, \Sigma_2^{-1} (\mu_{2 \, | \dataset} - \mu_2)~, \label{eq:backward_mean} \\
\text{Var}[X_1 | \dataset] 
    &= \frac{\nu +  \sum_{i=1}^{M^{l+1}}\left(\frac{(\mu_{2 | \dataset} - \mu_2)^2 + \Sigma_{2|\dataset}}{\Sigma_2}\right)_i}{\nu + d_2 - 2} \, \Sigma_{1 | 2} \nonumber \\
    &+ \Sigma_{12} \, \Sigma_2^{-1} \Sigma_{2|\dataset} \, \Sigma_2^{-1} \Sigma_{12}\T~,\label{eq:backward_cov}\\
\nu_{X_1 | \dataset} &= \nu_{X_1} + 1~, \label{eq:backward_dof} 
\end{align}
respectively, with $\Sigma_{1 | 2} = \Sigma_1 - \Sigma_{12}  \, \Sigma_2^{-1}  \, \Sigma_{21}$ and $\sum_{i=1}^{M^{l+1}}(.)_i$ being the sum over all elements of the vector.
For the first step (updating $a_l$ by $z_{l+1}$), the cross-scale matrix  $\Sigma_{12}$ is equal to $\Sigma_{a_l z_{l+1}}$ and can be calculated according to
\begin{align}
    \Sigma_{a_l z_{l+1}} 
    = \frac{\nu_{z_l}}{\nu_{z_l} - 2}\sigma_{z_{l+1}}^2 + \mu_{z_{l+1}}^2 - \mu_{a_l} \, \mu_{z_{l+1}}~. \label{eq:Sigma_az}
\end{align}

In the second step, when updating $\left(W_l\T, z_l\T\right)\T$ using $a_{l}$, the cross-covariance matrix (not cross-scale matrix) $C_{\left(W_l\T, z_l\T\right)\T a_l}$ has already been derived by \citet{KBNN}.
It is given by 
\begin{align}
    C_{W_l z_l a_l} =
\begin{bmatrix}
\operatorname{diag} \big(C_{W_l}^1 \cdot \mu_{z_l}, \dots, C_{W_l}^{M_l} \cdot \mu_{z_l} \big) \\
C_{z_l} \cdot \mu_{W_l}^1 \cdots C_{z_l} \cdot \mu_{W_l}^{M_l}
\end{bmatrix}. \label{eq:Sigma_Cwza}
\end{align}
The scale matrix can then be obtained by rescaling the covariance matrix according to Equation~\eqref{eq:scale_matr}.

In Alg.~\ref{alg:HABNN} HABNN is summarized. First, the parameters of the predictive distribution $p(y|x)$ according to~\eqref{eq:predictive_distr} are calculated in lines 2--5. 
The output value $y$ is used to update the weights and thus, to calculate the weight posterior $p(W|x,y)$ according to~\eqref{eq:posterior}. This is performed in lines 6--10.
Applying HABNN sequentially over all input-output pairs $(x_i, y_i) \in \dataset$, $i=1,\ldots,N$ in a single run completes the training of the BNN.

\begin{algorithm}[t]
   \caption{Single training pass for HABNN}
   \label{alg:HABNN}
\begin{algorithmic}[1]
   \STATE {\bfseries Input:} data instance $(x_i, y_i)$
   \FOR{$l=1$ {\bfseries to} $L$}
   \STATE Calculate $(\mu_a)_{i}^{l} \text{ and } (\sigma_a^2)_{i}^{l}$ by Eq. \eqref{eq:mean_a} and \eqref{eq:cov_a} $ \quad \forall i$
   \STATE Calculate $(\mu_z)_{i}^{l} \text{ and } (\sigma_z^2)_{i}^{l}$  by Eq. \eqref{eq:exp_val_f(a)} and \eqref{eq:cov_f(a)} $ \quad \forall i$
   \ENDFOR
    \FOR{$l=L$ {\bfseries to} $1$}
    \STATE Update $\mu_{a|D}^{l}, \Sigma_{a| \dataset}^{l} \text{ and } \nu_{a|\dataset}$ by Eq. \eqref{eq:backward_mean}--\eqref{eq:Sigma_az}
    \STATE Update $\mu_{W|\dataset}^{l}, \Sigma_{W|\dataset}^{l} \text{ and } \nu_{W|\dataset}$ by Eq.~\eqref{eq:backward_mean}--\eqref{eq:backward_dof}, \eqref{eq:Sigma_Cwza}
    \STATE Update $\mu_{z|\dataset}^{l}, \Sigma_{z|\dataset}^{l} \text{ and } \nu_{z|\dataset}$ by Eq.~\eqref{eq:backward_mean}--\eqref{eq:backward_dof}, \eqref{eq:Sigma_Cwza}
    \ENDFOR
\end{algorithmic}
\end{algorithm}

\subsection{Comparing the Backward Pass with TAGI and KBNN}
Both KBNN and TAGI share the update formulas
\begin{align}
    \mathbb{E}[X_1 \, | \, \dataset] 
    &= \mu_{1|\dataset} \nonumber \\
    &=  \mu_1 + \Sigma_{12} \, \Sigma_2^{-1} (\mu_{2 \, | \dataset} - \mu_2)~, \label{eq:KBNN_mean_update} \\
    \text{Var}[X_1\,|\,\dataset] 
    &= \Sigma_{1 | \dataset} \nonumber \\
    &= \Sigma_{1 | 2} + \Sigma_{12} \, \Sigma_2^{-1} \Sigma_{2|\dataset} \, \Sigma_2^{-1} \Sigma_{12}\T \label{eq:KBNN_cov_update}
\end{align}
for the mean and covariance matrix, respectively. 
Here, the mean update step \eqref{eq:KBNN_mean_update} is identical to the mean update~\eqref{eq:backward_mean} of HABNN. 

\newpage

The covariance matrix update \eqref{eq:KBNN_cov_update} is also similar to the scale matrix update~\eqref{eq:backward_cov} in HABNN, with one key difference: a scaling factor 
\[
\frac{\nu + \sum_{i=1}^{M^{l+1}}\left(\frac{(\mu_{2 | \dataset} - \mu_2)^2 + \Sigma_{2 | \dataset}}{\Sigma_2}\right)_i}{\nu + d_2 - 2}
\]
precedes \( \Sigma_{1 | 2} \) in HABNN.
However, as the degrees of freedom $\nu$ approach infinity with an increasing number of training samples, this scaling factor converges to~1.

Consequently, in the limit of large degrees of freedom, the scale matrix update in HABNN becomes equivalent to the covariance update in TAGI and KBNN.
As the Student's t-distribution converges to a Gaussian one, the scale matrix of HABNN transitions to a covariance matrix, aligning both update steps fully. 

Thus, HABNN generalizes both TAGI and KBNN, preserving their methodologies while extending their applicability through a more flexible prior structure.

Since HABNN employs a mean‐field variational approximation over its weight vector of size \(n\) (i.e.\ the total number of individual network weights), it needs only to learn \(2n + 1\) parameters in total: the mean and variance (or scale) for each of the \(n\) weights, plus a single global degrees‐of‐freedom parameter. 
Consequently, both its memory footprint and computational cost scale linearly in \(n\). 
As shown in Table~\ref{tab: training time test set} (Appendix), this lean parameterization also makes HABNN the fastest model among those evaluated.

\section{Experiments}
\label{sec:experiments}
In this section, we will evaluate HABNN by comparing its performance to other state-of-the-art methods on UCI regression datasets as this is the standard benchmark for BNN models. 
This can be seen in the papers of for example \citeauthor{KBNN, TAGI, PBP} and many more.
Additionally, we will analyze how effectively HABNN can handle out-of-distribution data and how trustworthy its predictions are.

For all experiments, we will use a neural network with one hidden layer of size 50 and ReLU activation.
The only exception is Laplace which was implemented with Tanh as its ReLU implementation diverged most of the time on most datasets (approx. 90\% of the time).

The initial scale matrix for HABNN will be set as the identity matrix times 0.01, while the means will be initialized by drawing from a standard Gaussian distribution, following a similar approach to that used in \cite{KBNN}.
The initial degrees of freedom for the weights $W$ will  be set to 12.

All experiments have been conducted on a Intel Core i9-13950HX on a Lenovo P16 Gen2 laptop.

\subsection{Experiments on UCI Regression Datasets}
\label{sec:experiments_uci}
Six UCI datasets will be used to compare the performance of the different models: Concrete, Energy, Wine, Naval, Yacht, and Kin8nm. 
The datasets will be randomly split into training and test sets, with 90\% of the data allocated for training and 10\% for testing. 
Importantly, we will not perform any data preprocessing, as this can lead to overfitting, as previously discussed in Section~\ref{sec: Problem Formulation} and further shown in the appendix.
We will later demonstrate that avoiding data preprocessing can significantly enhance the models' out-of-distribution performance. 
This is particularly important in high-security environments, where BNNs were designed to ensure reliable performance even when the underlying data-generating distribution may shift.
For SVI, MCMC the implementation in Pyro \citep{pyro} is used, for PBP we use the Theano implementation of the original paper and \citet{backpack_laplace} for Laplace.

We train KBNN, TAGI, HABNN, PBP, and Dropout for one epoch each whereas SVI is trained for 250 epochs since it is used to batch over the whole training set.
For MCMC, we employ the No-U-Turn sampler and run sampling until 100 posterior draws are obtained, while for the Laplace approximation we perform one epoch of optimization to fit the point estimates of the weights and a second epoch to compute their variances.

In order to account for stochastic effects, we average the performance of the models on 100 runs and state their median RMSE and negative log-likelihood (NLL) as well as their OOD performance measured by \eqref{Eq: nll_ood} and \eqref{Eq: rmes_ood}, respectively.
Remember that the OOD scenarios have been simulated by multiplying the test set by a factor of 0.1 and 2, or by adding three times the standard deviation of the test set.
Note that the standard deviations are reported in Appendix~\ref{sec: std_dev}.

\begin{table*}[t]
\captionsetup{justification=centering}
\caption{Average Root Mean Square Error Relative Percentage (Actual RMSE Values) \\ Ranking Legend: \textbf{Best Performing}, \secondbest{Second Best}, \thirdbest{Third Best}}
\centering
\small
\setlength{\tabcolsep}{3pt}
\begin{tabular}{lcccccccc}
\toprule
Dataset & PBP & Dropout & SVI & KBNN & TAGI & HABNN & MCMC & Laplace \\
\midrule
Concrete & 19 (692.55) & 93 (21.68) & \thirdbest{18} (21.73) & 7100 (82.33) & 157 \textbf{(13.36)} & \secondbest{15} \secondbest{(16.82)} & 2384 (20.13) & \textbf{0} \thirdbest{(16.96)}\\
Energy & \thirdbest{26} (241.23) & 36 (17.65) & \secondbest{15} (12.83) & 6760 (19.53) & 314 \secondbest{(5.98)} & 67 \thirdbest{(9.62)} & 3494 \textbf{(3.74)} & \textbf{1} (9.71) \\
Wine & \textbf{20} (5.18) & 335 (0.98) & \secondbest{41} (1.33) & 554 (1.97) & 571 \textbf{(0.77)} & \thirdbest{50} \thirdbest{(0.93)} & 458 \secondbest{(0.83)} & 13 (0.97) \\
Naval & \textbf{0} \secondbest{(0.02)} & \thirdbest{30} (48.05) & \secondbest{5} (267.19) & 31855 (104.60) & 53467 \textbf{(0.01)} & 197 (0.12) & 227 (3353) & 33 \thirdbest{(0.06)}\\
Yacht & 31 (205.31) & 11 (14.71) & 60 \secondbest{(10.45)} & 28 \thirdbest{(14.51)} & \thirdbest{10} (14.55) & \secondbest{4} (15.28) & 289 \textbf{(8.80)} & \textbf{3} (15.79) \\
Kin8nm & \textbf{1} (0.30) & 156 (0.16) & 276 \textbf{(0.11)} & 231 (0.16) & \thirdbest{103} \secondbest{(0.13)} & \secondbest{56} (0.49) & 593 (8.80) & 113 \thirdbest{(0.15)}\\
\bottomrule
Average & \textbf{16} (190.77) & 110 (17.21) & 69 (52.27) & 7755 (37.18) & 9104 \textbf{(5.80)} & \thirdbest{65} \secondbest{(7.23)} & 1241 (564.43) & \secondbest{27} \thirdbest{(7.26)}
\end{tabular}
\label{tab: RMSE results}
\end{table*}

\begin{table*}[t]
\captionsetup{justification=centering}
\caption{Average Negative Log-Likelihood Relative Percentage (Actual NLL Values) \\ Ranking Legend: \textbf{Best Performing}, \secondbest{Second Best}, \thirdbest{Third Best}}
\centering
\small
\setlength{\tabcolsep}{3pt}
\begin{tabular}{lcccccccc}
\toprule
Dataset & PBP & Dropout & SVI & KBNN & TAGI & HABNN & MCMC & Laplace \\
\midrule
Concrete & \secondbest{3} (11.43) & 238 (23410) & \textbf{0} (5.63) & \thirdbest{6} (8.69) & 703 \secondbest{(4.95)} & 13 \thirdbest{(5.59)} & 161 \textbf{(4.39)} & 69 (17909) \\
Energy & \thirdbest{11} (21.64) & 88 (15558) & \secondbest{2} (5.38) & \textbf{0} (8.24) & 582 \thirdbest{(4.80)} & 40 \secondbest{(3.81)} & 1158 \textbf{(2.84)} & 68 (4324) \\
Wine & \thirdbest{13} (1353) & 1992 (46.42) & \textbf{5} \thirdbest{(3.51)} & \secondbest{6} (4.97) & 4829 (5.86) & 29 \secondbest{(2.89)} & 545 \textbf{(1.21)} & 50 (247.9) \\
Naval & $>10^6$ ($>10^6$) & \thirdbest{33} ($>10^6$) & \textbf{6} (8.27) & \secondbest{10} (9.68) & 68640 \secondbest{(-0.20)} & 65 \thirdbest{(2.88)} & $>10^6$ ($>10^6$) & 196 \textbf{(-1.43)}\\
Yacht & \textbf{14} (7.06) & 24 (10807) & \secondbest{18} (3.77) & \thirdbest{20} \thirdbest{(4.37)} & 20305 (5.84) & 27 \secondbest{(4.27)} & 46 \textbf{(3.56)} & 95 (2171) \\
Kin8nm & 260 (275.95) & 11821 (-0.08) & \thirdbest{388} \secondbest{(-0.59)} & 821 (-0.41) & \secondbest{128} (-0.45) & \textbf{22} (2.87) & 767 \textbf{(-1.03)} & 552 \thirdbest{(-0.43)} \\
\bottomrule
Average & $>10^6$ ($>10^6$) & 2366 ($>10^6$) & \secondbest{70} \thirdbest{(4.33)} & \thirdbest{144} (5.92) & 15864 \textbf{(3.47)} & \textbf{33} \secondbest{(3.72)} & $>10^6$ ($>10^6$) & 172 (4108)
\end{tabular}
\label{tab: NLL results}
\end{table*}

Tables~\ref{tab: RMSE results} and~\ref{tab: NLL results} clearly illustrate that HABNN is the only model to secure a top‐three ranking on every dataset across both RMSE and NLL metrics, reflecting its unmatched consistency and robustness. 
In terms of RMSE, HABNN achieves 1st on Concrete (15 \%), 2nd on Energy (67 \%), 2nd on Wine (50 \%), 3rd on Naval (197 \%), 2nd on Yacht (4 \%), and 1st on Kin8nm (56 \%), yielding an average relative RMSE error of 7.23 \%—second only to PBP’s 7.26 \%. 
For NLL, HABNN similarly attains 3rd on Concrete (13 \%), 2nd on Energy (40 \%), 2nd on Wine (29 \%), 2nd on Naval (65 \%), 3rd on Yacht (27 \%), and 1st on Kin8nm (22 \%), resulting in an average NLL error of 3.72 \%, again ranking second overall. 

\newpage

No other method matches this level of across‐the‐board performance: PBP and SVI excel on some datasets but falter on others, TAGI and KBNN depend heavily on data normalization and degrade under extreme scales, MCMC and Laplace exhibit instability or variance collapse, and Dropout manages at most a single top‐three finish across both metrics. 
HABNN’s heavy‐tailed multivariate Student’s~$t$ prior fosters extensive exploration of the parameter space while its local Gaussian‐like updates ensure precise convergence to high‐probability posterior modes, thereby combining strong uncertainty quantification with competitive predictive accuracy in a way no other model achieves.  
This is further confirmed by the work of \citet{Priors_revisited}, as they showed that BNNs with fully connected layers exhibit heavy-tailed weight distributions.

\subsection{Industrial Benchmark}
To assess HABNN’s practical utility, we evaluate it on the \emph{Industrial Benchmark} (IB), a continuous‐state, continuous‐action, partially‐observable environment explicitly designed to capture core challenges of real industrial control: delayed responses, rate‐of‐change limits on actuators, heteroscedastic sensor noise, latent drift, and multi‐criteria reward trade‐offs \citep{SIB}. 

\vspace{1cm}
Unlike classic reinforcement learning (RL) testbeds (e.g.\ CartPole, MountainCar), IB embeds realistic actuator constraints and noisy sensor observations that scale with latent state, forcing policies to remain adaptive over long horizons rather than converge to a fixed operating point \citep{SIB}.  
A more detailed explanation about the benchmark as well as the experimental setup can be found in Appendix~\ref{sec: add_exp_indstr_bench_}.
  
Recent work introduced a Model Predictive Control + Active Learning framework that uses online uncertainty estimates to sample next states where the model is most uncertain, achieving large gains in sample efficiency on IB \citep{xinyang_uncertainty}. 
Adopting a similar uncertainty‐aware strategy, we compare HABNN to standard RL baselines—A2C, TD3, and SAC—from Stable Baselines \citep{stable-baselines}.

\begin{figure}[t]
  \centering
  \begin{tikzpicture}
    \begin{axis}[
        width=3.1in, 
        height=2.5in, 
        xlabel={Timestep},
        ylabel={Loss},
        xmin=10, xmax=60,
        ymin=300, ymax=1100,
        xtick={10,20,30,40,50,60},
        ytick={300,500,700,900,1100},
        grid=both,
        legend style={
          at={(0.5,-0.25)}, 
          anchor=north,
          legend columns=2,
          /tikz/every even column/.append style={column sep=0.5cm},
          draw=none, 
          font=\small
        },
        cycle list={ 
          solid, thick,
          densely dashed, thick,
          dotted, thick,
          dashdotted, thick
        },
        tick label style={font=\small},
        label style={font=\small},
        title style={font=\small}
      ]
      \addplot+[solid, mark=*, mark options={solid}] table[x=Timestep,y=HABNN] {
        Timestep HABNN A2C TD3 SAC
        10 335 381 318 405
        20 381 458 409 491
        30 376 487 576 625
        40 357 512 794 805
        50 401 556 911 1027
        60 352 570 909 1075
      };
      \addlegendentry{HABNN}

      \addplot+[densely dashed, mark=square*, mark options={solid}] table[x=Timestep,y=A2C] {
        Timestep HABNN A2C TD3 SAC
        10 335 381 318 405
        20 381 458 409 491
        30 376 487 576 625
        40 357 512 794 805
        50 401 556 911 1027
        60 352 570 909 1075
      };
      \addlegendentry{A2C}

      \addplot+[dotted, mark=triangle*, mark options={solid}] table[x=Timestep,y=TD3] {
        Timestep HABNN A2C TD3 SAC
        10 335 381 318 405
        20 381 458 409 491
        30 376 487 576 625
        40 357 512 794 805
        50 401 556 911 1027
        60 352 570 909 1075
      };
      \addlegendentry{TD3}

      \addplot+[dashdotted, mark=diamond*, mark options={solid}] table[x=Timestep,y=SAC] {
        Timestep HABNN A2C TD3 SAC
        10 335 381 318 405
        20 381 458 409 491
        30 376 487 576 625
        40 357 512 794 805
        50 401 556 911 1027
        60 352 570 909 1075
      };
      \addlegendentry{SAC}

    \end{axis}
  \end{tikzpicture}
  \caption{Performance on the Industrial Benchmark: loss over timesteps for HABNN, A2C, TD3, and SAC. Lower is better.}
  \label{fig:ib_performance}
\end{figure}
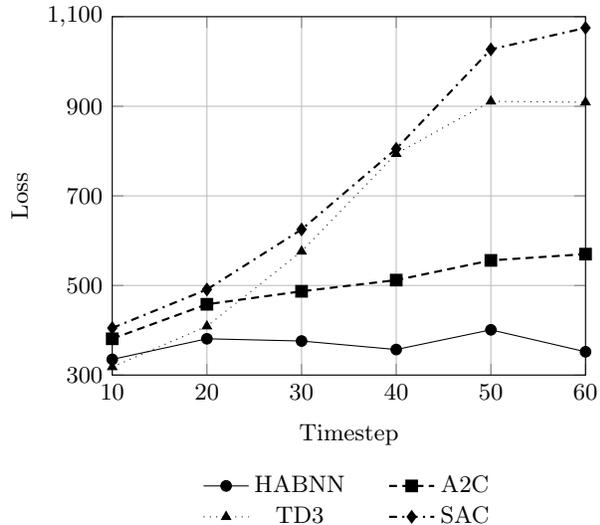

Figure~\ref{fig:ib_performance} shows that HABNN achieves the lowest and most stable one‐step prediction loss on IB compared to standard baselines (A2C, TD3, SAC). From the first timesteps, HABNN drives its mean‐squared error into the 300–400 range and holds it there throughout training, whereas the baselines start above 380 and either plateau at higher loss levels or exhibit sharp spikes under IB’s noisy, non‐stationary dynamics. 

\vspace{1cm}
Both TD3 and SAC struggle with heteroscedastic sensor noise and actuator delays—spiking in loss as they attempt to average out disturbances—while A2C converges more slowly and never matches HABNN’s low‐loss regime. 

These results confirm that HABNN’s uncertainty‐aware sampling, powered by heavy‐tailed Student’s~$t$ priors and a latent‐drift model, yields far more sample‐efficient and robust system identification, achieving and sustaining low prediction error with orders of magnitude fewer interactions.

\section{Limitations}
HABNN’s tractability relies on two main approximations. In the forward pass, we fit a Student’s \(t\) distribution—ignoring alternative families and higher moments—a strategy common to most non‐sampling BNNs. 
In the backward pass, we assume that pre‐activations (\(a_l\)), post‐activations (\(z_l\)), weights, and previous‐layer inputs (\(z_{l-1}\)) are jointly multivariate Student’s \(t\), which admits closed‐form updates but may misrepresent complex posteriors. 
Performance can degrade if a high‐probability mode lies near the initialization; in practice, increasing the degrees of freedom \(\nu\) restores Gaussian behavior.
For very deep networks, the heavy tails at initialization may still induce instability, suggesting that layer‐wise \(\nu\) schedules or adaptive update rates could be useful.

\section{Conclusion and Future Work}
We presented the Hierarchical Approximate Bayesian Neural Network (HABNN), which offers analytical forward and backward passes with linear complexity in the number of weights. 

HABNN consistently matches or outperforms state‐of‐the‐art BNN methods, demonstrates strong robustness to out‐of‐distribution inputs, and maintains stability across initializations. 
It supports true online learning, trains faster than competing approaches (cf.\ Tab.~\ref{tab: training time test set}), and eliminates the need for a learning‐rate hyperparameter, thus simplifying deployment and reducing tuning effort. 
These characteristics make HABNN particularly well-suited for safety‐critical applications where reliable uncertainty quantification and minimal preprocessing are essential. 
Future work will extend HABNN to classification tasks, enhance scalability in deeper architectures, and explore richer posterior approximations to further narrow the gap to fully sampling‐based methods.

\bibliographystyle{plainnat}
\bibliography{aistats2026.bib}

\clearpage
\appendix
\thispagestyle{empty}

\onecolumn
\aistatstitle{Supplementary Material}

In the supplementary material the following additional derivations and experiments can be found:
\begin{enumerate}[label=\textbf{Section \Alph*}, leftmargin =*, wide=0pt]
    \item Prior distribution of the BNN's weights.
    \item A detailed derivation of HABNN's forward pass.
    \item A detailed derivation of HABNN's backward pass.
    \item Comparison of the training times of different BNN algorithms. 
    \item Contains an additional evaluation of HABNN on different initializations of the scale parameter, the degrees of freedom and amount of hidden layers.
    \item Full experiments on non-normalized UCI regression datasets.
    \item Full experiments on normalized UCI regression datasets.
    \item Standard deviations of the most important experiments.
    \item Lists additional details about the industrial benchmark as well as the experimental setup used.
\end{enumerate}

\section{Weight Distribution}
\label{sec:Appendix Weight Distribution}
Most BNNs model the weights as Gaussian distributed \citep{SVI, KBNN, VI}.
This is, however, worrying as a Gaussian distribution puts a lot of probability mass around its mean due to its exponential decay. 
This in turn makes it challenging for the model to effectively explore its parameter space in depth.
As a result, the models become highly sensitive to their initializations and struggle to fit the data effectively if the posterior mode is far from the initialization.

To effectively address this issue, we will introduce a hierarchical distribution for the weights.
We will add a Gaussian-inverse-gamma distribution on top of the Gaussian distribution of the weights. 
This means that the weights will be distributed according to
\begin{align*}
    W \, | \, \mu_W, \sigma_W^2 &\sim \mathbb{N}(\mu_W, \sigma_W^2)~, \\
    (\mu_W, \sigma_W^2) &\sim \mathbb{N-}\Gamma^{-1}(\mu_{\mu}~, \lambda, \alpha, \beta)~,
\end{align*}
where the latter Gaussian-inverse-gamma distribution is defined as follows:

\begin{itemize}
    \item \( \mu_W \) represents the mean of the weight distribution and, given the variance \( \sigma_W^2 \), follows
    \[
    \mu_W \, | \, \sigma_W^2 \sim \mathbb{N}\left(\mu_{\mu}, \frac{\sigma_W^2}{\lambda}\right)~.
    \]
    \item \( \sigma_W^2 \) follows an inverse-gamma distribution with shape parameter \( \alpha \) and scale parameter \( \beta \) according to  
    \[
    \sigma_W^2 \sim \Gamma^{-1}(\alpha, \beta)~.
    \]
    This distribution is a common prior for variance in Bayesian inference, particularly when the variance is uncertain.
\end{itemize}
It can now easily be shown that the resulting distribution of the weights will be a Student's $t$-distribution \citep{ConjugatePrior}.
The Student's $t$-distribution is characterized by heavier tails than the normal distribution, making it more robust to outliers. 
It is commonly used to model data with greater variability or when dealing with small sample sizes.
As the degrees of freedom \( \nu \) increases, the Student’s $t$-distribution approaches a normal distribution.
This is one of the fundamental properties of the Student's $t$-distribution, allowing it to be particularly useful in Bayesian statistics as it can model the data with robustness to extreme values in the early stages of training.

As layers of (Bayesian) neural networks typically consist of more than one weight due to multiple inputs to a layer, we will assume a joint Gaussian distribution with a Gaussian-inverse-Wishart hyperprior distribution.

This, in turn, will result in multivariate Student's $t$-distributions for the weights, which is defined by the probability density function (PDF)
\begin{align*}
f(x \, | \, \mu, \Sigma, \nu) 
&= 
\frac{\Gamma\left(\frac{\nu + p}{2}\right)}{\Gamma\left(\frac{\nu}{2}\right) (\nu \pi)^{p/2} |\Sigma|^{1/2}} \\
&\cdot \left[ 1 + \frac{1}{\nu} (x - \mu)^\top \Sigma^{-1} (x - \mu) \right]^{-\frac{\nu + p}{2}},
\end{align*}
where $\mu$ is the location, $\Sigma$ the scale matrix, and $\nu$ represents the degrees of freedom.

As the Student's $t$-distribution does not belong to the Exponential family, there exists no conjugate prior for this distribution. Thus, there is no analytical solution to Equations~\eqref{eq:bayes_posterior} and \eqref{eq:predictive_distr}. Instead we follow an assumed density approach where we assume a (multivariate) Student's $t$-distribution for the output $z^l$ of any given layer and for the posterior distribution of the weights $W^l$ of that layer.

As weights in a given layer of a BNN become independent over time, it is common to assume their independence from the outset. 
This approach, known as the \emph{mean-field assumption}, simplifies computations by treating all weights as uncorrelated, effectively setting non-diagonal elements of the covariance matrices to zero.
Adopting this assumption not only increases scalability but also aligns with observations made by \citet{TAGI}, who demonstrated that the covariance matrix of the posterior tends to converge toward a diagonal form.

\section{Derivation Forward Pass}
\label{sec:Appendix Forward Pass}
The PDF of the Student’s $t$-distribution with \( \nu \) degrees of freedom, location parameter \( \mu \), and scale parameter \( \sigma^2 \) is given by

\[
f(x \, | \, \mu, \sigma^2, \nu) = \frac{\Gamma\left(\frac{\nu+1}{2}\right)}{\sqrt{\nu \pi} \, \Gamma\left(\frac{\nu}{2}\right) \sigma} \left( 1 + \frac{(x - \mu)^2}{\nu \, \sigma^2} \right)^{-\frac{\nu+1}{2}}
\]
where \( \Gamma(\cdot) \) is the gamma function.
The multivariate $t$-distribution is defined by the PDF
\begin{align*}
    f(x \, | \, \mu, \Sigma, \nu) &= 
\frac{\Gamma\left(\frac{\nu + p}{2}\right)}{\Gamma\left(\frac{\nu}{2}\right) (\nu \pi)^{p/2} |\Sigma|^{1/2}} \\
&\cdot \left[ 1 + \frac{1}{\nu} (x - \mu)^\top \Sigma^{-1} (x - \mu) \right]^{-\frac{\nu + p}{2}},
\end{align*}
where $\mu$ is the location, $\Sigma$ the scale matrix, and $\nu$ represents the degrees of freedom.
Note that for the remainder of the text we will also refer to the location (parameter) as mean.

Given that $a$ follows a Student's $t$-distribution, we assume the parameters are defined by $\mu$, $\tau^2$, and $\nu$, representing the mean, scale, and degrees of freedom, respectively.
We will omit the quantity $D$ to ease the notation.
The mean $\mu_z$ of the post-activation is then given by
\begin{align*}
    &\mathbb{E}[ \, \mathrm{f}(a) ] 
= \int_{-\infty}^{\infty} \mathrm{f}(a) \, \mathrm{p}(a) \, \dd a \\
&= \frac{\Gamma(\frac{\nu+1}{2})}{\Gamma(\frac{\nu}{2}) \sqrt{\pi \nu \tau^2}} \,\int_0^\infty a \cdot \left(1 + \frac{(a-\mu)^2}{\nu \tau^2} \right)^{\frac{-(\nu+1)} {2}} \dd a~.
\end{align*}

Substituting $x = a - \mu$ results in 
\begin{align}
    \mathbb{E}[ \, \mathrm{f}(a) ] &=\frac{\Gamma(\frac{\nu+1}{2})}{\Gamma(\frac{\nu}{2}) \sqrt{\pi \nu \tau^2}} \, \int_{-\mu}^{\infty} x \cdot \left(1 + \frac{x^2}{\nu \tau^2}\right)^{\frac{-(\nu+1)}{2}} \dd x \nonumber \\
    &+ \mu \, \underbrace{\frac{\Gamma(\frac{\nu+1}{2})}{\Gamma(\frac{v}{2}) \sqrt{\pi \nu \tau^2}} \int_{-\mu}^{\infty} \left(1 + \frac{x^2}{\nu \tau^2}\right)^{\frac{-(\nu+1)}{2}} \dd x}_{=\,1 - F(0 \, | \, \mu,\tau^2, \nu)}\label{eq:exp_val_int_1}
\end{align}
with $F(0 \, | \, \mu,\tau^2, \nu)$ being the cumulative distribution function of the random variable $a$.
This can easily be seen by simply resubstituting $a = x + \mu$ into the integral.

The first integral in \eqref{eq:exp_val_int_1} can now be approached similarly to calculating the expected value of a Student's $t$-distributed random variable. 
By substituting $u = 1 + \frac{x^2}{\nu \tau^2}$ one obtains 
\begin{align}
    \int_{-\mu}^{\infty} x \cdot \left(1 + \frac{x^2}{\nu \tau^2}\right)^{\frac{-(\nu+1)}{2}} \dd x &= \frac{\nu \tau^2}{2} \int_{1 + \frac{\mu^2}{\nu \tau^2}}^{\infty} u^{\frac{-(\nu+1)}{2}} \dd u \nonumber \\
    &= \frac{\nu \tau^2}{\nu-1} \left(1 + \frac{\mu^2}{\nu \tau^2}\right)^{\frac{1-\nu}{2}}. \label{eq:exp_val_int_2}
\end{align}
Combining Equation \eqref{eq:exp_val_int_1} and Equation \eqref{eq:exp_val_int_2} results in the final formula 
\begin{align}
    \mathbb{E}[\, \mathrm{f}(a)] &= \frac{\Gamma(\frac{\nu+1}{2})}{\Gamma(\frac{\nu}{2}) \sqrt{\pi \nu \tau^2}} \, \frac{\nu \tau^2}{\nu-1} \left(1 + \frac{\mu^2}{\nu \tau^2}\right)^{\frac{1-\nu}{2}} \nonumber \\
    &+ \mu \, (1 - \text{F}(\, 0\, | \, \mu,\tau^2, \nu)) =: \mu_z~. \label{eq:final_exp_val_f(a)}
\end{align}

This sums up the calculation of the mean and we will turn our attention towards the scale matrix in the following. 

The most convenient way to calculate the scale parameter is by actually calculating the variance and then rescaling it according to Equation \eqref{eq:scale_matr}.
Therefore, we need to compute the variance $\sigma_z^2$ of $f(a)$ next.
With the mean already derived, it is sufficient to merely calculate the (non-central) second-order moment $\mathbb{E}[\, \mathrm{f} (a)^2]$, which is equal to $\sigma_z^2 + \mu_z^2$. Reformulating the latter then gives the desired variance~$\sigma_z^2$.

To simplify the notation, we will temporarily omit the scaling factor of the density and include it later.
Hence, one obtains
\[
\mathbb{E}[\mathrm{f}(a)^2] \propto \int_0^\infty a^2 \cdot \left(1 + \frac{(a-\mu)^2}{\nu \tau^2} \right)^{\frac{-(\nu+1)} {2}} \dd a~,
\]
which can now be evaluated by substituting $x = a - \mu$ again, resulting in
\[
= \int_{-\mu}^{\infty} (x+\mu)^2 \left(1 + \frac{x^2}{\nu \tau^2}\right)^{\frac{-(\nu+1)}{2}} \dd x~.
\]

\newpage

After expanding the binomial term one obtains
\begin{align*}
    &= \! \underbrace{\int_{-\mu}^{\infty} x^2 \left(1 + \frac{x^2}{\nu \tau^2}\right)^{\frac{-(\nu+1)}{2}} \dd x}_{\textbf{(I)}}  \\
&+  
2 \mu \! \underbrace{\int_{-\mu}^{\infty} x \left(1 + \frac{x^2}{\nu \tau^2}\right)^{\frac{-(\nu+1)}{2}} \dd x}_{\textbf{(II)}} \, \\
&+ \, \mu^2 \! \! \underbrace{\int_{-\mu}^{\infty} \left(1 + \frac{x^2}{\nu \tau^2}\right)^{\frac{-(\nu+1)}{2}} \dd x}_{\textbf{(III)}}
\end{align*}
where the integrals in \textbf{II} and \textbf{III} have already been calculated in Equation~\eqref{eq:exp_val_int_2} and Equation~\eqref{eq:exp_val_int_1}, respectively.

Thus, the only one remaining is \textbf{I}. 
It can be handled by partitioning the domain of integration according to
\begin{align}
&\int_{-\mu}^{\infty} x^2 \left(1 + \frac{x^2}{\nu \tau^2}\right)^{\frac{-(\nu+1)}{2}} \dd x \\
&= \int_{0}^{\infty} x^2 \left(1 + \frac{x^2}{\nu \tau^2}\right)^{\frac{-(\nu+1)}{2}} \dd x \nonumber \\
&+ \int_{-\mu}^{0} x^2 \left(1 + \frac{x^2}{\nu \tau^2}\right)^{\frac{-(\nu+1)}{2}} \dd x. \label{eq:var_int}
\end{align}
The first integral can now be solved similarly to calculating the variance of a Student's t-distributed random variable.
By splitting the two terms in the integral, one obtains
\begin{align*}
&\int_{0}^{\infty} x^2 \left(1 + \frac{x^2}{\nu \tau^2}\right)^{\frac{-(\nu+1)}{2}} \dd x \\
&= \int_{0}^{\infty} \frac{x}{\left(1 + \frac{x^2}{\nu \tau^2}\right)^{\frac{1}{2}}} \frac{1}{\left(1 + \frac{x^2}{\nu \tau^2}\right)^{\frac{\nu-4}{2}}} \frac{x}{\left(1 + \frac{x^2}{\nu \tau^2}\right)^{2}} \,  \dd x~.
\end{align*}
Substituting $u = \frac{x^2}{\nu \tau^2 + x^2}$ results in 
\begin{align}
\frac{(\nu \tau^2)^{\frac{3}{2}}}{2} \, \int_{0}^{1} u^{\frac{1}{2}} \, (1 - u)^{\frac{\nu-4}{2}} \, \dd u = 
\frac{(\nu \tau^2)^{\frac{3}{2}}}{2} \hspace{0.15cm} \textbf{B}\left(\frac{3}{2}, \frac{\nu-2}{2}\right) \label{eq:substitution}   
\end{align}
with \textbf{B}$(\cdot)$ being the Beta function.
Note that we use the first definition of the Beta function, which is defined as 
\begin{align}
    \textbf{B}(x, y) = \int_0^1 u^{x-1} (1 - u)^{y-1} \, \dd u~. \label{eq:beta_fct}
\end{align}

Exploiting the close relationship between the Beta function and the Gamma function yields
\begin{align*}
\frac{(\nu \tau^2)^{\frac{3}{2}}}{2} \hspace{0.15cm} \textbf{B}\left(\frac{3}{2}, \frac{\nu-2}{2}\right) 
&= \frac{(\nu \tau^2)^{\frac{3}{2}}}{2} \, \frac{\Gamma\left(\frac{3}{2}\right) \, \Gamma\left(\frac{\nu-2}{2}\right)}{\Gamma\left(\frac{\nu+1}{2}\right)} \\
&= \frac{(\nu \tau^2)^{\frac{3}{2}}}{(\nu-2)} \frac{\frac{\sqrt{\pi}}{2} \, \Gamma\left(\frac{\nu}{2}\right)}{\Gamma\left(\frac{\nu+1}{2}\right)}~, 
\end{align*}
where we used the fact that $\Gamma\left(\frac{k-2}{2}\right) = \frac{\Gamma\left(\frac{k}{2}\right)}{\frac{k-2}{2}}$\,. 

Now, all that is left is to calculate the second summand in Equation \eqref{eq:var_int}.
As both cases can be calculated similarly, w.l.o.g we will assume $\mu \geq 0$.
One can split up the two terms in the integral just like before and use the same substitution as in Equation \eqref{eq:substitution} to get 
\begin{align*}
    &\int_{-\mu}^{0} x^2 \left(1 + \frac{x^2}{\nu \tau^2}\right)^{\frac{-(\nu+1)}{2}} \dd x \\
    &= \frac{(\nu \tau^2)^{\frac{3}{2}}}{2} \, \int_{0}^{\frac{\mu^2}{\nu \tau^2 + \mu^2}} x^{\frac{1}{2}} \, (1 - x)^{\frac{\nu-4}{2}} \, \dd x~. 
\end{align*}
The last term can now be recognized as the definition of the incomplete beta function, denoted by $\textbf {B}_{\frac{\mu^2}{\nu \tau^2 + \mu^2}}\left(\frac{3}{2}, \frac{\nu-2}{2}\right) $\,.  
Unlike the complete beta function, the incomplete beta function evaluates the integral only up to the argument, rather than over the entire range.
Hence, it is defined as
\begin{align*}
    \textbf{B}_{z}(x, y) = \int_0^{z} u^{x-1} (1 - u)^{y-1} \, \dd u.
\end{align*}

Combining all of the terms and simplifying them now results in 
\begin{align}
\mathbb{E}[\, \mathrm{f}(a)^2] 
&=\frac{\nu \tau^2}{2 (\nu-2)} \nonumber \\
&+ \mathrm i \hspace{0.15cm} \frac{\nu \tau^2}{2 \sqrt{\pi}} \hspace{0.15cm} \frac{\Gamma(\frac{\nu+1}{2})}{\Gamma(\frac{\nu}{2})}  \hspace{0.15cm} \textbf {B}_{\frac{\mu^2}{\nu \tau^2 + \mu^2}}\left(\frac{3}{2}, \frac{\nu-2}{2}\right) \nonumber \\
&+ 2 \mu \, (\mu_y - \mu \ (1 - \text{F}(\, 0 \, | \, \mu,\tau^2, \nu))) \nonumber \\
&+ \mu^2 \, \text{F}(\, \mu \, | \, 0,\tau^2, \nu)~, \label{eq:final_fa2}
\end{align}
where i is defined as 
\[
\mathrm i = 
\begin{cases} 
    1 & \text{if } \mu \geq 0 \\
    -1 & \text{otherwise }
\end{cases}~.
\]
Combining \eqref{eq:final_exp_val_f(a)} with \eqref{eq:final_fa2} yields the variance of $\mathrm{f}(a)$ according to
\begin{align}
    \text{Var}[\, \mathrm{f}(a)] &= \mathbb{E}[\, \mathrm{f}(a)^2] - \mu_z^2 \nonumber \\
    &= \frac{\nu \tau^2}{2 (\nu-2)} \nonumber \\
    &+ \mathrm i \hspace{0.15cm} \frac{\nu \tau^2}{2 \sqrt{\pi}} \hspace{0.15cm} \frac{\Gamma(\frac{\nu+1}{2})}{\Gamma(\frac{\nu}{2})}  \hspace{0.15cm} \textbf{B}_{\frac{\mu^2}{\nu \tau^2 + \mu^2}}\left(\frac{3}{2}, \frac{\nu-2}{2}\right) \nonumber \\
&+ 2 \mu \, (\mu_y - \mu \ (1 - \text{F}(\, 0 \, | \, \mu,\tau^2, \nu))) \nonumber \\
&+ \mu^2 \, \text{F}(\, \mu \, | \, 0,\tau^2, \nu) - \mu_z^2 \nonumber \\
&=: \sigma_z^2~.\label{eq:final_cov_f(a)}
\end{align}
Recall that this is not the final scale parameter but rather the variance, which needs to be rescaled according to Equation \eqref{eq:scale_matr} in order to obtain the actual scale parameter.

We assume that the degrees of freedom remain unaltered as the ReLU function is mostly linear and linear transformations do not change the degrees of freedom of a Student's t-distributed random variable.

For the output layer $l=L$, the variance $\sigma_\epsilon^2$ of the Gaussian noise $\epsilon$ is added to the variance $\sigma_z^2$ of \eqref{eq:final_cov_f(a)} to account for the aleatoric uncertainty of the data.

\newpage
\section{Derivation Backward Pass}
\label{sec: appendix Backward Pass}

Central to the approach presented here is the conditional distribution of a partitioned multivariate t-distribution. 

Let X follow a multivariate t-distribution with the partition 
\[
X = \begin{pmatrix} X_1 \\ X_2 \end{pmatrix} \sim 
\text{t}_{d}\left(\begin{pmatrix} \mu_1 \\ \mu_2 \end{pmatrix}, \begin{pmatrix} \Sigma_1, \Sigma_{12} \\ \Sigma_{21}, \Sigma_{2} \end{pmatrix}, \nu \right)~,
\]
then the conditional distribution $X_1 \, | \, X_2$ is given by
\begin{align}
    X_1 \, | \, X_2 \sim 
    \text{t}_{d_1} \left( \mu_{1 | 2}, 
    \widetilde{\Sigma}_{1 | 2}, \,
    \nu + d_2 \right) \label{eq:cond_distr}
\end{align}
where 
\begin{align*}
    \widetilde{\Sigma}_{1 | 2} &= \frac{\nu + (X_2 - \mu_2)\T \Sigma_2^{-1}(X_2 - \mu_2)}{\nu + d_2} \, \Sigma_{1 | 2} \\ \Sigma_{1 | 2} &= \Sigma_1 - \Sigma_{12}  \, \Sigma_2^{-1}  \, \Sigma_{21}, \\
    \mu_{1 | 2} &= \mu_1 + \Sigma_{12} \Sigma_{2}^{-1} (X_2 - \mu_2),
\end{align*}
and $d_i$ being the dimension of $X_i$ holds.

In the first scenario, we aim to update the pre-activation $a_l$ using $z_{l+1}$.
The cross-scale matrix  $\Sigma_{a_l z_{l+1}}$ can be calculated according to
\begin{align}
    \Sigma_{a_l z_{l+1}} 
    &= \mathbb{E}[a_l \, z_{l+1}] - \mathbb{E}[a_l] \, \mathbb{E}[z_{l+1}] \nonumber \\  
    &= \mathbb{E}[a_l \, \mathrm{f}(a_l)] - \mu_{a_l} \, \mu_{z_{l+1}} \nonumber \\
    &= \int_{-\infty}^{\infty} a_l \, \mathrm{f}(a_l) \, \mathrm{p}(a_l) \dd a_l - \mu_{a_l} \, \mu_{z_{l+1}}  \nonumber \\
    &= \int_{0}^{\infty} a_l \, a_l \, \mathrm{p}(a_l) \dd a_l - \mu_{a_l} \, \mu_{z_{l+1}} \nonumber \\
    &= \int_{-\infty}^{\infty} \mathrm{f}(a_l)^2 \, \mathrm{p}(a_l) \dd a_l - \mu_{a_l} \, \mu_{z_{l+1}} \nonumber \\
    &= \mathbb{E}[\, \mathrm{f}(a_l)^2] - \mu_{a_l} \, \mu_{z_{l+1}} \nonumber \\
    &= \frac{\nu_{z_l}}{\nu_{z_l} - 2}\sigma_{z_{l+1}}^2 + \mu_{z_{l+1}}^2 - \mu_{a_l} \, \mu_{z_{l+1}}~. 
\end{align}
In the second case, when updating $\left(W_l\T, z_l\T\right)\T$ using $a_{l}$, the cross-covariance matrix (not cross-scale matrix) $\Sigma_{\left(W_l\T, z_l\T\right)\T a_l}$ has already been derived by \citet{KBNN}.
It is given by 
\begin{align}
    C_{W_l z_l a_l} &=
\begin{bmatrix}
C_{W_l}^1 \cdot \mu_{z_l} & \cdots & 0 \\
\vdots & \ddots & \vdots \\
0 & \cdots & C_{W_l}^{M_l} \cdot \mu_{z_l} \\
C_{z_l} \cdot \mu_{W_l}^1 & \cdots & C_{z_l} \cdot \mu_{W_l}^{M_l}
\end{bmatrix} \nonumber \\
&=
\begin{bmatrix}
\operatorname{diag} \big(C_{W_l}^1 \cdot \mu_{z_l}, \dots, C_{W_l}^{M_l} \cdot \mu_{z_l} \big) \\
C_{z_l} \cdot \mu_{W_l}^1 \cdots C_{z_l} \cdot \mu_{W_l}^{M_l}
\end{bmatrix}. 
\end{align}
The scale matrix can then be obtained by rescaling the covariance matrix according to Equation~\eqref{eq:scale_matr}.
Attentive readers will notice that $\mu_{1 | 2}$ and $\Sigma_{1 | 2}$ are the parameters of a conditional distribution of a multivariate Gaussian distribution.
As the degrees of freedom increase (by $d_2$ for every sample), the scaling factor of the scale matrix will converge to one.
As a multivariate Student's $t$-distribution converges to a multivariate Gaussian distribution as the degrees of freedom increase \citep{multv_t_dstr_cnv_multv_gauss}, the conditional distribution of a multivariate Student's t-distribution similarly approaches that of a multivariate Gaussian distribution.

Since KBNN and TAGI use a comparable training framework with Gaussian distributions, the update steps in HABNN will likewise converge to those of KBNN and TAGI. 
This can also be formally demonstrated once the final backward propagation formulas have been derived.

To calculate the posterior of $X_1$, we marginalize over $X_2$, yielding 
\begin{align}
    p(X_1 | \dataset) &= \int_{\Omega_{X_2}} p(X_1 \, | \, X_2) \, p(X_2 \, | \, \dataset) \, \dd X_2~.  \label{eq:posterior_appendix}
\end{align}
Unlike the Gaussian case, where the posterior \eqref{eq:posterior_appendix} can be computed analytically \citep{KBNN}, this is not possible here because the product of two multivariate t probability density functions (PDFs) does not yield another multivariate t PDF.
To address this issue, we approximate the true posterior in Equation \eqref{eq:posterior_appendix} by using a surrogate multivariate t-distribution, matching its moments to those of the true posterior. 
In both cases considered, we will assume a joint multivariate t-distribution for all of the random variables.

This means that we will assume that $X_1 := a_l$ and $X_2 := z_{l+1}$ are jointly multivariate t-distributed in the first case and $X_1:= \left( W_{l}\T, z_{l}\T \right)\T$ and $X_2 := a_{l}$ are also jointly multivariate t-distributed in the second case.
This implies that the moments of the posterior can be derived in a general form using the random variables $X_1$ and $X_2$.
Subsequently, the update formula can be adjusted based on the specific parameters being updated at any given point.

In both cases considered, we will assume a joint multivariate t-distribution for all of the random variables.
This means that we will assume that $X_1 := a_l$ and $X_2 := z_{l+1}$ are jointly multivariate t-distributed in the first case and $X_1:= \left( W_{l}\T, z_{l}\T \right)\T$ and $X_2 := a_{l}$ are also jointly multivariate t-distributed in the second case.
This implies that the moments of the posterior can be derived in a general form using the random variables $X_1$ and $X_2$.
Subsequently, the update formula can be adjusted based on the specific parameters being updated at any given point.  \par
\par
We will start off by calculation the mean, which is given as 

\begin{align}
&\mathbb{E}[X_1 \, | \, \dataset] 
= \int_{\Omega_{X_1}} X_1 \, p(X_1 \, | \, \dataset) \, \dd X_1  \nonumber \\
&\stackrel{\text{a)}}{=} \int_{\Omega_{X_1}} X_1 \left(\int_{\Omega_{X_2}} p(X_1 \, | \, X_2) \hspace{0.15cm}  p(X_2 \, | \, \dataset) \, \dd X_2 \right) \, \dd X_1  \nonumber \\
&\stackrel{\text{b)}}{=} \int_{\Omega_{X_2}} \left(\int_{\Omega_{X_1}} X_1 \, p(X_1 \, | \, X_2) \hspace{0.15cm} \dd X_1 \right) p(X_2 \, | \, \dataset) \hspace{0.15cm} \dd X_2~, 
\label{eq:marginal-mean}
\end{align}
where a) has been obtained by marginalizing the posterior density over $X_2$ and b) by switching the orders of integration.
This can be done as the conditions for Tonelli's theorem \citep{tonelli_thrm} for non-negative measurable functions are obviously met. 
Recall that the parameters of a conditional multivariate t-distribution are given by Equation \eqref{eq:cond_distr}. 

\newpage
In order to enhance readability, we will denote the parameters $\mu_{X_i}$, $\Sigma_{X_i}$ by $\mu_i$, $\Sigma_i$ and we will not adjust $\dataset$ in the formulae.
One then gets

\begin{align}
&\mathbb{E}[X_1 \, | \, \dataset] 
= \int_{\Omega_{X_1}} X_1 \, p(X_1 \, | \, \dataset) \hspace{0.15cm} \dd X_1 \nonumber \\
&\stackrel{\eqref{eq:marginal-mean}}{=} \int_{\Omega_{X_2}} \left(\int_{\Omega_{X_1}} X_1 \hspace{0.1cm} p(X_1 \, | \, X_2) \hspace{0.15cm} \dd X_1 \right) p(X_2 \, | \, \dataset) \hspace{0.15cm} \dd X_2 \nonumber \\
&= \int_{\Omega_{X_2}} \left( \mu_1 + \Sigma_{12} \, \Sigma_2^{-1} (X_2 - \mu_2) \right) \, p(X_2 \, | \, \dataset) \hspace{0.15cm} \dd X_2 \nonumber \\
&= \mu_1 + \Sigma_{12} \, \Sigma_2^{-1} \underbrace{\int_{\Omega_{X_2}} X_2 \hspace{0.1cm} p(X_2 \, | \dataset) \hspace{0.15cm} \dd X_2}_{=\,\mu_{2 \, | \, D}} - \Sigma_{12} \, \Sigma_2^{-1} \mu_2 \nonumber \\
&= \mu_1 + \Sigma_{12} \, \Sigma_2^{-1} (\mu_{2 \, | \dataset} - \mu_2)~. \label{eq:final_backward_mean}
\end{align}

The sets $\Omega_{X_1}, \Omega_{X_2}$ denote the sample spaces of $X_1$ and $X_2$, respectively.
Attentive readers will note that this is the same update step as in \citet{KBNN}. 
Next, $\mathbb{E}[X_1 X_1\T \, | \, D]$ will be calculated as it is central for the calculation of the covariance matrix of the weights under the posterior.

The approach is similar to the one used for the mean, as the posterior distribution will be marginalized again and then the updated parameters of $X_2$ will be inserted.
Remember that earlier we decided to disregard any non-diagonal entries of the covariance matrices and scale matrices, respectively, as they not only diminish over time but also significantly degrade the scalability. This then leads to
\begin{align}
&\mathbb{E}[X_1 X_1\T \, | \, \dataset]  
= \int_{\Omega_{X_1}} X_1 X_1\T \, p(X_1 | \dataset) \hspace{0.15cm} \dd X_1 \nonumber \\
&\stackrel{\text{a)}}{=} \int_{\Omega_{X_2}} \left(\int_{\Omega_{X_1}} X_1 X_1\T \, p(X_1 \, | \, X_2) \dd X_1 \right) \, p(X_2 \, | \, \dataset) \hspace{0.15cm} \dd X_2 \nonumber \\
&\stackrel{\text{b)}}{=} \underbrace{\int_{\Omega_{X_2}} \frac{\nu + d_2}{\nu + d_2 - 2} \, \widetilde{\Sigma}_{1 | 2}\, \, p(X_2 \, | \, \dataset) \hspace{0.15cm} \dd X_2}_{\mathbf{I}} \nonumber \\
&+ \underbrace{\int_{\Omega_{X_2}} \mu_{1 | 2} \, \, \mu_{1 | 2}\T \, \,  p(X_2 \, | \, \dataset) \hspace{0.15cm} \dd X_2}_{\mathbf{II}}. \label{eq:var_posterior}
\end{align}

Here, for a) we marginalized over $X_2$ and used Tonelli's theorem again and for b) we inserted the formula $\mathbb{E}[X_2 X_2\T \, | \, \dataset]~=~\text{Var}(X_2 \, | \, \dataset)~+~\mu_{2 \, | \, \dataset} \, \mu_{2 \, | \, \dataset}\T$.
When looking at $\mathbf{I}$ one gets
\begin{align}
    \int_{\Omega_{X_2}} \frac{\nu + d_2}{\nu + d_2 - 2} \, \widetilde{\Sigma}_{1 | 2} \, \, p(X_2 \, | \, \dataset) \hspace{0.15cm} \dd X_2 \nonumber \\
   = \frac{\nu +  \sum_{i=1}^{M^{l+1}} \left( \frac{(\mu_{2 \, | \dataset} - \mu_2)^2 + \Sigma_{2 \, | \, \dataset}}{\Sigma_2} \right)_i}{\nu + d_2 - 2} \, \Sigma_{1 | 2} \label{eq:posterior_sub_var_1}
\end{align}
where the division of $\Sigma_2$ is meant element-wise and $\sum_i$ is the sum over all elements of the vector. 
Note that we took advantage of the diagonal covariance structure here. 

The second term $\mathbf{II}$ in \eqref{eq:posterior_sub_var_1} can now be calculated by simply multiplicating out the terms, rearranging them and then solving the trivial integrals.
By doing so one obtains
\begin{align}
    &\int_{\Omega_{X_2}} \mu_{1 | 2} \, \mu_{1 | 2}\T \, p(X_2 \, | \, \dataset) \hspace{0.15cm} \dd X_2  \nonumber\\
    =&\ \mu_1 \, \mu_1\T \nonumber \\
    &+ \underbrace{\int_{\Omega_{X_2}} \mu_1 (X_2 - \mu_2)\T \, (\Sigma_2^{-1})\T \, \Sigma_{12}\T \, p(X_2 \, | \, \dataset) \hspace{0.15cm} \dd X_2}_{=\mu_1 (\mu_{2 \, |\dataset} - \mu_2)\T \, (\Sigma_2^{-1})\T  \, \Sigma_{12}\T} \nonumber \\ 
    &+ \underbrace{\int_{\Omega_{X_2}} \Sigma_{12} \, \Sigma_2^{-1} (X_2 - \mu_2)\T \mu_1^{T} \, p(X_2 \, | \, \dataset) \hspace{0.15cm} \dd X_2}_{= \Sigma_{12} \, \Sigma_2^{-1} (\mu_{2 \, |\dataset} - \mu_2) \, \mu_1\T} \nonumber \\
    &+ \Sigma_{12} \, \Sigma_2^{-1} \left(\Sigma_{2|\dataset} \, + \, \left\| \mu_{2 | \dataset} - \mu_2 \right\|_2^2 \right) \, (\Sigma_2^{-1})\T \, \Sigma_{12}\T \nonumber \\[0.08cm]
    =&\ \Sigma_{12} \, \Sigma_2^{-1} \Sigma_{2|\dataset} \, \Sigma_2^{-1} \Sigma_{12}\T + \mu_{1|\dataset} \, \mu_{1|\dataset}\T~. \label{eq:posterior_sub_var_2}
\end{align}
Combining the Equations \eqref{eq:posterior_sub_var_1} and \eqref{eq:posterior_sub_var_2} now allows solving the integral in Equation \eqref{eq:var_posterior}.
In conclusion this leads to 
\begin{align}
    \text{Var}[X_1 | \dataset] &= \mathbb{E}[X_1 X_1\T \, | \, \dataset] - \mu_{1|\dataset} \, \mu_{1|\dataset}\T \nonumber \\
    &= \frac{\nu +  \sum_{i=1}^{M^{l+1}} \left( \frac{(\mu_{2 \, | \dataset} - \mu_2)^2 + \Sigma_{2 \, | \, \dataset}}{\Sigma_2} \right)_i}{\nu + d_2 - 2} \, \Sigma_{1 | 2} \nonumber \\
    &+ \Sigma_{12} \, \Sigma_2^{-1} \Sigma_{2|\dataset} \, \Sigma_2^{-1} \Sigma_{12}\T~. \nonumber
\end{align}
To obtain the scale matrix, being the second parameter of the multivariate t-distribution, the covariance matrix must be rescaled again as shown in Equation \eqref{eq:scale_matr}. 
Since the degrees of freedom of $X_1 | X$ are unknown, one must match moments again to calculate them analytically.
Calculating the fourth moment however involves complex and tedious computations.
A more efficient approach is to simply assume $\nu_{1 | D} = \nu + d_2$, simplifying the determination of the scale matrix to
\begin{align}
    \Sigma_{1| \dataset} &= \frac{\nu +  \sum_{i=1}^{M^{l+1}} \left( \frac{(\mu_{2 \, | \dataset} - \mu_2)^2 + \Sigma_{2 \, | \, \dataset}}{\Sigma_2} \right)_i}{\nu + d_2} \, \Sigma_{1 | 2} \nonumber \\
    &+ \Sigma_{12} \, \Sigma_2^{-1} \Sigma_{2|\dataset} \, \Sigma_2^{-1} \Sigma_{12}\T~. \label{eq:final_backward_cov}
\end{align}
In our case $d_2 = 1$ holds for both cases of $X_2$.
Combining all the formulas derived in the Section \ref{sec: Foward Pass} and \ref{sec: Backward Pass} completes the HABNN approach for calculating the predictive distribution and for updating the posterior weight distribution given training data $\dataset$. 

In total, the mean, scale matrix, and degrees of freedom of the pre-activation $a$ in the $l$-th layer will be updated according to Equation~\eqref{eq:final_backward_mean}, ~\eqref{eq:final_backward_cov}, and~\eqref{eq:cond_distr}, respectively, leading to
\begin{align}
    \mu_{a_l | \dataset} &= \mu_{a_l} + \Sigma_{a_l z_{l+1}} \, \Sigma_{z_{l+1}}^{-1} (\mu_{z_{l+1} \, | \dataset} - \mu_{z_{l+1}})~, \label{eq:a_final_backward_mean} \\
    \Sigma_{a_l| \dataset} &= \frac{\nu_{a_l}}{\nu_{a_l} + 1} \Sigma_{a_l | z_{l+1}} \nonumber \\
    &+ \frac{\sum_{i=1}^{M^{l+1}}\left(\frac{(\mu_{z_{l+1} | \dataset} - \mu_{z_{l+1}})^2 + \Sigma_{z_{l+1}|\dataset}}{\Sigma_{z_{l+1}}}\right)_i}{\nu_{a_l} + 1} \Sigma_{a_l | z_{l+1}} \nonumber \\
    &+ \Sigma_{a_l z_{l+1}} \, \Sigma_{z_{l+1}}^{-1} \Sigma_{z_{l+1}|\dataset} \, \Sigma_{z_{l+1}}^{-1} \Sigma_{a_l z_{l+1}}\T~, \label{eq:a_final_backward_cov} \\
    \nu_{a_l | \dataset} &= \nu_{a_l} + 1~. \label{eq:a_final_backward_dof}
\end{align}
Similary, the update steps of the weights $W$ are given by 
\begin{align}
    \mu_{W_l | \dataset} &= \mu_{W_l} + \Sigma_{W_l a_{l}} \, \Sigma_{a_{l}}^{-1} (\mu_{a_{l} \, | \dataset} - \mu_{a_{l}})~, \label{eq:W_final_backward_mean} \\
    \Sigma_{W_l| \dataset} &= \frac{\nu_{W_l}}{\nu_{W_l} + 1} \Sigma_{W_l | a_{l}} \nonumber \\
    &+ \frac{\sum_{i=1}^{M^{l+1}}\left(\frac{(\mu_{z_{l+1} | \dataset} - \mu_{z_{l+1}})^2 + \Sigma_{z_{l+1}|\dataset}}{\Sigma_{z_{l+1}}}\right)_i}{\nu_{W_l} + 1} \Sigma_{W_l | a_{l}} \nonumber \\
    &+ \Sigma_{W_l a_{l}} \, \Sigma_{a_{l}}^{-1} \Sigma_{a_{l}|\dataset} \, \Sigma_{a_{l}}^{-1} \Sigma_{W_l a_{l}}\T~, \label{eq:W_final_backward_cov} \\
    \nu_{W_l | \dataset} &= \nu_{W_l} + 1~. \label{eq:W_final_backward_dof}
\end{align}
And the ones of the post-activation $z$ by
\begin{align}
    \mu_{z_l | \dataset} &= \mu_{z_l} + \Sigma_{z_l a_{l}} \, \Sigma_{a_{l}}^{-1} (\mu_{a_{l} \, | \dataset} - \mu_{a_{l}})~, \label{eq:z_final_backward_mean} \\
    \Sigma_{z_l| \dataset} &= \frac{\nu_{z_l}}{\nu_{z_l} + 1} \Sigma_{z_l | a_{l}} \nonumber \\
    &+ \frac{\sum_{i=1}^{M^{l+1}}\left(\frac{(\mu_{z_{l+1} | \dataset} - \mu_{z_{l+1}})^2 + \Sigma_{z_{l+1}|\dataset}}{\Sigma_{z_{l+1}}}\right)_i}{\nu_{z_l} + 1} \Sigma_{z_l | a_{l}} \nonumber \\
    &+ \Sigma_{z_l a_{l}} \, \Sigma_{a_{l}}^{-1} \Sigma_{a_{l}|\dataset} \, \Sigma_{a_{l}}^{-1} \Sigma_{z_l a_{l}}\T~, \label{eq:z_final_backward_cov} \\
    \nu_{z_l | \dataset} &= \nu_{z_l} + 1~. \label{eq:z_final_backward_dof}
\end{align}
For $l=L$, this backward recursion commences from $\mu_{z_{l+1}} = y$ and $\Sigma_{z_{l+1}} = 0$\,.

\section{Training times}
\label{sec: Training times}
As we assess the performance of these models, it is crucial to consider training times. 
Given that different models utilize varying batch sizes, we will calculate each model's training time by iterating through the dataset sample by sample, updating the model after each pass. 
However, this approach presents a challenge for PBP, as its implementation requires a minimum of two samples for training. 
Consequently, we chose to iterate through the entire dataset using a batch size of 2, without any additional adjustments to the training time.

It's important to note that the training times for TAGI are excluded from this analysis because its current implementation is not compatible with Windows, necessitating the use of virtual machines. 
This limitation significantly impacts its performance, making a fair comparison impossible.

\begin{table*}[t]
\captionsetup{justification=centering}
\caption{Training time in seconds for different UCI regression datasets.}
\centering
\Large
\setlength{\tabcolsep}{6pt}
\begin{tabular}{lcccccccc}
\toprule
Dataset & PBP    & Dropout & SVI    & KBNN  & HABNN & MCMC & Laplace \\
\midrule
Concrete & 719.93 & 40.29   & 724.20 & 1.38   & 1.36  & $>10^3$   & 14.34   \\
Energy   & 537.75 & 27.87   & 516.45 & 0.95   & 0.88  & $>10^3$   & 184.92      \\
Wine     & 4482   & 5091    & 4410   & 9.03   & 7.60  & $>10^3$   & 1714      \\
Naval    & 8135   & 422.47  & 7807   & 15.08  & 12.08 & $>10^3$   & 5526      \\
Yacht    & 236.62 & 12.14   & 227.57 & 0.43   & 0.39  & $>10^3$   & 60.56      \\
Kin8nm   & 5527   & 297.61  & 5524   & 9.40   & 8.39  & $>10^3$   & 2167      \\
\bottomrule
\end{tabular}
\label{tab: training time test set}
\end{table*}

As shown in Table \ref{tab: training time test set}, KBNN and HABNN exhibit remarkable speed and significantly outperform all other models, thanks to their analytical expressions for both the forward and backward passes, which eliminate the need for costly gradient calculations. 
Although HABNN requires more complex computations for its forward pass, it remains faster than KBNN due to its efficient diagonal covariance structure.

It is quite revealing to note how slow PBP can be. 
Interestingly, it operates efficiently when it processes the entire training set at once, likely due to its backend being implemented in C++, which significantly accelerates the training process.
In this scenario, the training times (in seconds) for the datasets are as follows: Concrete (2.07), Energy (1.71), Wine (2.99), Naval (3.822), Yacht (1.61), and Kin8nm (2.99).
However, once we account for this effect by iterating through the training set sample by sample, its performance drastically declines.

\section{Weight Initialization and Architectures}
\label{sec:experiments_initialization}
Unlike traditional neural networks, BNNs employ random variables as weights, typically doubling the number of learnable parameters. 
This raises the question of whether and how different weight initializations affect the model's final performance.
To explore this, we evaluated HABNN's sensitivity to weight and degrees of freedom initializations by conducting experiments with varying initial scale parameters and degrees of freedom for the weights.
In these experiments, we set the scale parameters to 0.01, 0.1, 1, 2, and 5 and the degrees of freedom to 12, 20, 30, 50, and 75, measuring their impact on model performance across different datasets.

\begin{table*}[t]
\captionsetup{justification=centering}
\caption{RMSE (NLL) of HABNN for different UCI regression datasets and scale parameter initializations.}
\centering
\Large
\setlength{\tabcolsep}{6pt}
\begin{tabular}{lcccccccc}
\toprule
Dataset & 0.01 & 0.1 & 1 & 2 & 5  \\
\midrule
Concrete & 16.82 (5.59) & 16.91 (5.34) & 16.9 (5.59) & 17.16 (5.74) & 18.16 (6.81) \\
Energy & 9.71 (3.81) & 11.9 (5.15) & 8.8 (5.78) & 13.98 (6.84) & 13.29 (6.86) \\
Wine & 0.93 (2.89) & 0.94 (2.38) & 0.94 (2.43) & 0.89 (2.56) & 0.88 (2.7)\\
Naval & 0.06 (2.88) & 0.04 (0.53) & 0.26 (0.68) & 3.81 (7.91) & 7.99 (8.62) \\
Yacht & 15.28 (4.27) & 15.41 (8.16) & 14.87 (7.7) & 14.56 (7.18) & 14.86 (6.62) \\
Kin8nm & 0.49 (2.87) & 0.22 (0.44) & 0.21 (1.0) & 0.21 (1.12) & 0.22 (1.27) \\
\bottomrule
\end{tabular}
\label{tab:varinitscaling}
\end{table*}

\begin{table*}[t]
\captionsetup{justification=centering}
\caption{RMSE (NLL) of HABNN for different UCI regression datasets and degrees of freedom initializations.}
\centering
\Large
\setlength{\tabcolsep}{6pt}
\begin{tabular}{lcccccccc}
\toprule
Dataset & 12 & 20 & 30 & 50 & 75  \\
\midrule
Concrete & 16.82 (5.59) & 16.89 (5.67) & 16.77 (5.63) & 16.96 (5.52) & 17.22 (5.45) \\
Energy & 9.71 (3.81) & 9.99 (5.33) & 9.85 (5.19) & 10.22 (5.0) & 10.63 (4.89) \\
Wine & 0.93 (2.89) & 1.11 (2.68) & 1.14 (2.56) & 1.16 (2.41) & 1.29 (2.39) \\
Naval & 0.06 (2.88) & 0.05 (0.56) & 0.06 (0.49) & 0.03 (0.44) & 0.04 (0.48)\\
Yacht & 15.28 (4.27) & 15.21 (7.8) & 15.87 (7.63) & 15.67 (7.25) & 15.03 (6.48) \\
Kin8nm & 0.49 (2.87) & 0.44 (1.34) & 0.44 (1.29) & 0.39 (1.18) & 0.44 (1.42)\\
\bottomrule
\end{tabular}
\label{tab:dofinitscaling}
\end{table*}

Tables~\ref{tab:varinitscaling} and ~\ref{tab:dofinitscaling} demonstrate that HABNN exhibits remarkable robustness across most datasets, with only slight variations in performance. 
This consistent stability further reinforces HABNN's status as a highly robust model capable of adapting to diverse conditions.

The primary exception is observed in the naval dataset when the initial scale parameter is significantly high. 
This behavior could be attributed to the model's enhanced exploration due to its heavier tails combined with the elevated initial scale, making it challenging to converge effectively in the presence of the naval dataset's high feature values.
Further investigation is required to validate and explore this hypothesis in greater detail.

We also evaluated HABNN on architectures with 1 to 4 hidden layers, each containing 50 neurons. The experiments were conducted with different degrees of freedom and varying initializations of the scale parameter.

\begin{table*}[t]
\captionsetup{justification=centering}
\caption{RMSE (NLL) of HABNN for different amount of hidden layers, each of size 50.}
\centering
\Large
\setlength{\tabcolsep}{6pt}
\begin{tabular}{lccccc}
\toprule
Dataset & 1 & 2 & 3 & 4 \\
\midrule
Concrete & 16.82 (5.59) & 15.98 (7.21) & 32.21 (10.09) & 37.18 (9.3) \\
Energy & 9.71 (3.81) & (16.64 (7.14) & 27.51 (9.87) & 31.32 (9.7) \\
Wine & 0.93 (2.89) & 16.29 (7.18) & 33.96 (10.14) & 65.74 (9.79) \\
Naval & 0.06 (2.88) & 15.79 (7.16) & 28.43 (9.64) & 78.64 (9.93) \\
Yacht & 15.28 (4.27) & 15.92 (7.17) & 27.89 (9.62) & 64.44 (9.84) \\
Kin8nm & 0.49 (2.87) & 16.06 (7.11) & 28.43 (10.17) & 21.66 (9.08) \\
\bottomrule
\end{tabular}
\label{tab:architectures}
\end{table*}

As shown in Table~\ref{tab:architectures}, the single-hidden-layer architecture performs best in most cases, except for the Concrete and Yacht datasets, where the two-layer architecture closely matches or even exceeds its performance. Notably, only the four-layer model exhibits a decline in performance, likely due to overfitting on these relatively simple datasets.

While adjusting the number of hidden neurons per layer might have improved results for deeper architectures, this was beyond the scope of our study, which primarily focuses on demonstrating the stability of HABNN.

\section{Experiments on non-normalized UCI Regression datasets}

Below you can find the full experiments that were used to derive the results in the main body of the paper.

\subsection{In-distribution non-normalized UCI Regression datasets}
\label{sec: in_distr_non_norm}

In this subsection we will look at the in-distribution performance of all of the BNNs considered. 
The training data has not been normalized and the testing data has not been rescaled or changed in any way.
The architecture and training methods are the same as in the main body of the paper.

\begin{table*}[t]
\captionsetup{justification=centering}
\caption{RMSE on the test set with non-normalized training data}
\centering
\Large
\setlength{\tabcolsep}{4pt}
\begin{tabular}{lcccccccc}
\toprule
Dataset & PBP & Dropout & SVI & KBNN & TAGI & HABNN & MCMC & Laplace\\
\midrule
Concrete & 692.55 & 21.68 & 21.73 & 82.33 & 13.36 & 16.82 & 20.13 & 16.96 \\
Energy & 241.23 & 17.65 & 12.83 & 19.53 & 5.98 & 9.71 & 3.74 & 9.62\\
Wine & 5.18 & 0.98 & 1.53 & 1.97& 0.77 & 0.93 & 0.83 & 0.97\\
Naval & 0.02 & 48.05 & 267.19 & 104.6 & 0.01 & 0.06 & 3353 & 0.08 \\
Yacht & 205.31 & 14.71 & 10.45 & 14.51 & 14.55 & 15.28 & 8.8 & 15.79 \\
Kin8nm & 0.3 & 0.16 & 0.12 & 0.16 & 0.13 & 0.49 & 0.09 & 0.15\\
\bottomrule
\end{tabular}
\end{table*}

\begin{table*}[t]
\captionsetup{justification=centering}
\caption{NLL on the test set with non-normalized training data}
\centering
\Large
\setlength{\tabcolsep}{4pt}
\begin{tabular}{lcccccccc}
\toprule
Dataset & PBP & Dropout & SVI & KBNN & TAGI & HABNN & MCMC & Laplace\\
\midrule
Concrete & 11.43 & 23,410 & 5.63 & 8.69 & 4.95 & 5.59 & 4.39 & 17909 \\
Energy & 21.64 & 15,558 & 5.38 & 8.24 & 4.8  & 3.81 & 2.84 & 4324 \\
Wine & 1353 & 46.42 & 3.51 & 4.97 & 5.86 & 2.89 & 1.21 & 247.9 \\
Naval & $>10^6$ & $>10^6$ & 8.27 & 9.68 & -0.2 & 2.88 & $>10^6$ & -1.43 \\
Yacht & 7.06 & 10,807 & 3.77 & 4.37 & 5.84 & 4.27 & 3.56 & 2171 \\
Kin8nm & 275.95 & -0.08 & -0.59 & -0.41 & -0.45 & 2.87 & -1.03 & -0.43\\
\bottomrule
\end{tabular}
\end{table*}

\subsection{Out-of-distribution non-normalized UCI Regression datasets}
\label{sec: ood_non_norm}

Here are the OOD experiments in which the data has not been normalized.
This means that we have not normalized the training data and have then looked at the OOD performance of all the BNNs considered. 
Remember that we looked at 3 OOD settings, namely rescaling the test set by a factor of 0.1 and 2 or by adding three times the standard deviation of the test set onto itself.
The results can be found in the Tables~\ref{tab:scaling2}, \ref{tab:scaling01} and \ref{tab:scaling_std_dev}.

\begin{table*}[t]
\captionsetup{justification=centering}
\caption{Out-of-distribution RMSE (NLL) for UCI regression datasets rescaled by a factor of 2 with non-normalized training data}
\centering
\small
\setlength{\tabcolsep}{0.8pt}
\begin{tabular}{lcccccccc}
\toprule
Dataset & PBP & Dropout & SVI & KBNN & TAGI & HABNN & MCMC & Laplace \\
\midrule
Concrete & 1366.89 (11.92) & 38.76 (75100) & 28.48 (5.84) & 174.68 (9.37) & 40.09 (5.74) & 18.93 (4.38) & 36.85 (11.68) & 17.17 (5.03) \\
Energy & 480.86 (19.61) & 26.41 (35073) & 20.37 (6.34) & 50.11 (8.91) & 27.13 (5.6) & 18.46 (5.23) & 24.63 (25.93) & 9.62 (3.92)\\
Wine & 9.72 (997.89) & 5.08 (1290.7) & 2.1 (3.72)& 7.89 (5.6) & 5.56 (266.35) & 1.29 (1.73) & 3.95 (8.03) & 1.16 (5.32) \\
Naval & 0.02 ($>10^6$) & 86.21 ($>10^6$) & 150.22 (8.03) & 260.23 (10.55) & 6.95 (341.47) & 0.52 (0.67) & 620.58 (6418) & 0.05 (2.38) \\
Yacht & 344.82 (7.34) & 16.81 (14112) & 11.76 (3.93) & 17.77 (4.81) & 15.31 (2016) & 15.54 (4.15) & 22.31 (4.56) & 16.14 (4.26) \\
Kin8nm & 0.28 (74.34) & 0.25 (1.64) & 0.45 (0.81) & 0.33 (0.58) & 0.27 (0.1) & 0.27 (0.37) & 0.35 (7.57) & 0.3 (0.06)\\
\bottomrule
\end{tabular}
\label{tab:scaling2}
\end{table*}

\begin{table*}[t]
\captionsetup{justification=centering}
\caption{Out-of-distribution RMSE (NLL) for UCI regression datasets rescaled by a factor of 0.1 with non-normalized training data}
\centering
\small
\setlength{\tabcolsep}{1pt}
\begin{tabular}{lcccccccc}
\toprule
Dataset & PBP & Dropout & SVI & KBNN & TAGI & HABNN & MCMC & Laplace \\
\midrule
Concrete & 67.38 (9.09)   & 35.38 (62562)   & 25.91 (5.75)   & 35.90 (6.37)   & 35.44 (105.09) & 19.51 (4.42)   & 110.49 (7.56)  & 17.39 (4.36)             \\
Energy   & 23.80 (20.25)  & 23.96 (28696)   &  9.58 (3.80)   & 25.25 (5.92)   & 23.90 (29.82)  & 20.34 (5.35)   & 30.43 (21.99)  & 9.5 (4.44)    \\
Wine     &  1.29 (1325.61)&  5.02 (1259.00) &  1.43 (3.31)   &  5.80 (3.43)   &  5.05 (306.67) &  1.20 (1.65)   & 4.94 (7.21)    & 1.03 (3.12)             \\
Naval    &  0.02 ($>10^6$)&  5.34 (1531.00) & 11.07 (5.19)   & 13.00 (7.46)   &  0.91 (9.71)   &  0.46 (1.49)   & 6974 ($>10^6$) & 0.15 (2.2)      \\
Yacht    & 37.75 (5.25)   & 17.02 (14489)   & 14.34 (4.13)   & 17.52 (6.12)   & 16.62 (861.81) & 14.58 (4.11)   & 27.06 (4.87)   & 15.68 (4.69)           \\
Kin8nm   & 0.32 (1771.84) &  0.26 (1.96)    &  0.29 (2.33)   & 0.27 (1.65)    &  0.27 (0.19)   &  0.26 (0.33)   & 0.36 (3.69)    & 0.27 (5.16)           \\
\bottomrule
\end{tabular}
\label{tab:scaling01}
\end{table*}

\begin{table*}[t]
\captionsetup{justification=centering}
\caption{Out-of-distribution RMSE (NLL) for UCI regression datasets rescaled by three times the standard deviation with non-normalized training data}
\centering
\small
\setlength{\tabcolsep}{0.6pt}
\begin{tabular}{lcccccccc}
\toprule
Dataset & PBP & Dropout & SVI & KBNN & TAGI & HABNN & MCMC & Laplace \\
\midrule
Concrete & 247.96 (12.26) & 51.22 ($>10^6$) & 22.75 (5.32) & 17572 (11.89) & 27.63 (8.39) & 19.83 (5.74) & 1353 (15.16) & 16.37 (16809) \\
Energy & 28.29 (17.81) & 21.84 (23825) & 14.33 (5.71) & 3944 (9.78) & 23.26 (62.8) & 9.99 (5.40) & 348.2 (59.29) & 9.46 (4182) \\
Wine & 1.48 (1223.00) & 2.70 (363.96) & 2.11 (4.00) & 24.95 (5.05) & 4.89 (293.42) & 1.70 (2.80) & 5 (8.18) & 1.1 (360.72) \\
Naval & 0.02 ($>10^6$) & 95.36 ($>10^6$) & 680.13 (10.10) & $>10^6$ (13.95) & 8.21 (60.06) & 0.09 (0.83) & 25274 ($>10^6$) & 0.04 (-0.45) \\
Yacht & 41.81 (5.62) & 15.21 (11556) & 24.00 (5.25) & 20.64 (4.82) & 16.09 (697.19) & 17.69 (7.99) & 53.35 (6.2) & 16.92 (309.74) \\
Kin8nm & 0.31 (1138.00) & 0.72 (24.53) & 0.50 (1.95) & 0.99 (6.64) & 0.25 (0.09) & 1.76 (6.03) & 1.16 (9.35) & 0.39 (0.61) \\
\bottomrule
\end{tabular}
\label{tab:scaling_std_dev}
\end{table*}

\section{Further Experiments on normalized UCI Regression datasets}
\label{sec:appendix}

In this section we will report the full experiments when the data has been normalized.
Again, the network architecture as well as the hyperparameters stayed the same as reported in the main body of the paper.
The results can be found in the Tables~\ref{tab:in_distr_norm}~-~\ref{tab:norm_scaling_std_dev}.

\begin{table*}[t]
\captionsetup{justification=centering}
\caption{In-distribution RMSE (NLL) for different normalized UCI regression datasets.}
\centering
\normalsize
\setlength{\tabcolsep}{2pt}
\begin{tabular}{lcccccccc}
\toprule
Dataset & PBP & Dropout & SVI & KBNN & TAGI & MCMC & Laplace \\
\midrule
Concrete & 8.67 (3.78) & 31.49 (49553) & 6.28 (3.26) & 7.41 (3.4) & 13.36 (4.95) & 5.02 (4.39) & 7.72 (17909) \\
Energy   & 4.84 (4.89) & 19.03 (18108)  & 2.84 (2.5)  & 3.41 (3.24) & 3.44 (2.74)  & 0.67 (2.84) & 3.2  (4324)  \\
Wine     & 0.71 (12.46)& 0.88 (37.59)   & 0.70 (1.06) & 0.74 (13.75)& 0.71 (6.54)  & 0.68 (1.21) & 0.85 (247.90)  \\
Naval    & 0.01 (1.86) & 0.03 (-1.35)   & 0.14 (0.81) & 0.04 (1.14) & 0.00 (0.19)  & 0.01 ($>10^6$) & 0.02 (-1.43) \\
Yacht    & 7.92 (3.27) & 17.55 (15399)  & 1.16 (1.72) & 5.24 (3.44) & 5.75 (3.37)  & 0.41 (3.56) & 11.68 (2171.47) \\
Kin8nm   & 0.16 (4.36) & 0.16 (-0.03)   & 0.11 (-0.66)& 0.16 (-0.38)& 0.12 (0.28)  & 0.08 (-1.03)& 0.19 (-0.43) \\
\bottomrule
\end{tabular}
\label{tab:in_distr_norm}
\end{table*}

\begin{table*}[t]
\captionsetup{justification=centering}
\caption{Out-of-distribution RMSE (NLL) for different normalized UCI regression datasets in which the test set has been rescaled by a factor of 2.}
\centering
\normalsize
\setlength{\tabcolsep}{2pt}
\begin{tabular}{lcccccccc}
\toprule
Dataset & PBP & Dropout & SVI & KBNN & TAGI & MCMC & Laplace \\
\midrule
Concrete & 34.89 (4.99) & 20.72 (21462) & 39.14 (5.12) & 75.03 (5.78) & 40.09 (5.74) & 98.15 (11.68) & 35.15 (5.03) \\
Energy   & 11.82 (3.87) &  8.04 (3234)  & 63.77 (7.66) & 64.16 (5.66) & 23.66 (4.62) & 69.17 (25.93) & 10.07 (3.92) \\
Wine     & 26.91 (5.26) & 169.20 ($>10^6$) & 100.28 (6.20) & 45.13 (5.26) & 64.50 (19.18) & 84.63 (8.03) &  7.64 (5.32) \\
Naval    & $>10^6$ ($>10^6$) & 18.36 (16846) & 127.40 (6.50) & 88.72 (6.18) & $>10^6$ (24.03) & 768.11 (6418) &  0.21 (2.38) \\
Yacht    & 21.39 (5.35) & 18.36 (16846) & 127.40 (6.50) & 88.72 (6.18) & 50.02 (6.00) & 213.99 (4.56) & 15.56 (4.26) \\
Kin8nm   &  0.27 (2.96) &  0.25 (1.74)  &  0.33 (0.59) &  0.33 (0.47) &  0.26 (5.61) &   0.34 (7.57) &  0.21 (0.06) \\
\bottomrule
\end{tabular}
\label{tab:norm_scaling2}
\end{table*}

\begin{table*}[t]
\captionsetup{justification=centering}
\caption{Out-of-distribution RMSE (NLL) for different normalized UCI regression datasets in which the test set has been rescaled by a factor of 0.1.}
\centering
\normalsize
\setlength{\tabcolsep}{1pt}
\begin{tabular}{lcccccccc}
\toprule
Dataset & PBP & Dropout & SVI & KBNN & TAGI & MCMC & Laplace \\
\midrule
Concrete & 34.23 (4.95) & 17.82 (15866) & 57.87 (6.44) & 49.62 (6.63) & 35.44 (105.09) & 308.82 (7.56) & 16.45 (4.36) \\
Energy   & 21.25 (5.98) & 16.20 (13122) & 88.94 (14.46) & 22.35 (5.27) & 20.27 (4.45)   &  57.02 (21.99) & 20.13 (4.44) \\
Wine     & 28.06 (5.22) &250.92 ($>10^6$) & 88.04 (6.58) &561.76 (8.26) & 48.41 (14.10)  & 202.79 (7.21) &  1.05 (3.12) \\
Naval    & $>10^6$ ($>10^6$) & 19.34 (18684) & 17.14 (5.16) & 39.75 (5.95) & $>10^6$ (23.91) & 961.75 ($>10^6$) & 0.07 (2.20) \\
Yacht    & 23.73 (5.28) & 16.77 (14054) & 17.14 (5.16) & 39.75 (5.95) & 41.94 (5.84)   &  72.07 (4.87) & 26.42 (4.69) \\
Kin8nm   &  0.26 (288.85) &  0.26 (1.98) &  0.32 (2.73) &  0.27 (1.50) &  0.27 (11.00)  &   0.36 (3.69) &  0.34 (5.16) \\
\bottomrule
\end{tabular}
\label{tab:norm_scaling01}
\end{table*}

\begin{table*}[t]
\captionsetup{justification=centering}
\caption{Out-of-distribution RMSE (NLL) for different normalized UCI regression datasets in which the test set has been rescaled by three times the standard deviation.}
\centering
\normalsize
\setlength{\tabcolsep}{1pt}
\begin{tabular}{lcccccccc}
\toprule
Dataset & PBP & Dropout & SVI & KBNN & TAGI & MCMC & Laplace \\
\midrule
Concrete & 41.85 (5.23) & 24.49 (29966) & 104.78 (44.35) & 108.57 (7.28) & 30.25 (4.85) & 51.31 (15.16) & 25.84 (16809) \\
Energy   & 20.22 (5.99) & 12.77 (8150)  & 25.48 (6.61)   & 50.08 (5.34)   & 20.40 (4.49) & 49.69 (59.29) &  5.28 (4182)  \\
Wine     & 24.59 (4.91) &  1.90 (179.65)&  0.97 (1.42)   &  1.32 (1.98)   & 47.12 (11.78)&  1.22 (8.18)  &  0.88 (360.72)  \\
Naval    & $>10^6$ ($>10^6$) & 0.16 (-0.03)&  0.55 (2.14)   &  3.08 (2.79)   & $>10^6$ (23.97)&  6.51 ($>10^6$) &  0.02 (-0.45)  \\
Yacht    & 26.12 (5.30) & 15.67 (12266) & 123.02 (133.27)& 68.88 (6.28)   & 35.18 (5.84) &158.03 (6.20) & 18.29 (309.74) \\
Kin8nm   &  0.25 (128.53)& 0.72 (24.64)&  0.54 (1.97)   &  1.07 (6.10)   &  0.25 (8.92) &  0.82 (9.35)  &  1.26 (0.61)   \\
\bottomrule
\end{tabular}
\label{tab:norm_scaling_std_dev}
\end{table*}

\section{Standard Deviations on all experiments}
\label{sec: std_dev}

Below you can find the respective standard deviations of the BNNs, for the experiments in which the training data has not been normalized. 
We will report the standard deviations of the actual RMSE (NLL) values, as well as the average relative RMSE (NLL) errors.
The results are reported in the Tables~\ref{tab:std_dev_1}~-~\ref{tab:std_dev_nll_rel_error}.

\begin{table*}[t]
\captionsetup{justification=centering}
\caption{Standard deviations of the RMSE for non-normalized training data}
\centering
\Large
\setlength{\tabcolsep}{5pt}
\begin{tabular}{lcccccccc}
\toprule
Dataset  & PBP    & Dropout & SVI    & KBNN  & TAGI  & HABNN & MCMC  & Laplace \\
\midrule
Concrete & 99.88  & 2.66    & 9.93   & 10.80 & 1.23  & 5.63  & 2.36  & 1.01     \\
Energy   & 12.54  & 7.58    & 2.45   & 5.34  & 0.47  & 2.68  & 0.87  & 0.18     \\
Wine     & 0.04   & 0.17    & 0.31   & 0.24  & 0.11  & 0.17  & 0.90  & 0.02     \\
Naval    & 0.00   & 38.34   & 150.65 & 21.85 & 0.00  & 0.12  & 1694  & 0.01     \\
Yacht    & 549.26 & 3.29    & 2.16   & 2.80  & 2.28  & 2.64  & 2.57  & 1.15     \\
Kin8nm   & 0.00   & 0.01    & 0.03   & 0.01  & 0.02  & 2.87  & 0.01  & 0.04     \\
\bottomrule
\end{tabular}
\label{tab:std_dev_1}
\end{table*}

\begin{table*}[t]
\captionsetup{justification=centering}
\caption{Standard deviations for NLL on non-normalized training data}
\centering
\Large
\setlength{\tabcolsep}{5pt}
\begin{tabular}{lcccccccc}
\toprule
Dataset  & PBP    & Dropout & SVI   & KBNN  & TAGI  & HABNN & MCMC           & Laplace       \\
\midrule
Concrete & 2.25   & 10763   & 0.45  & 0.51  & 0.83  & 0.86  & 0.12    & $>10^6$   \\
Energy   & 9.88   & 9308    & 0.46  & 0.56  & 0.52  & 0.26  & 0.22    & $>10^6$   \\
Wine     & 571.95 & 20.86   & 0.23  & 1.62  & 0.84  & 0.01  & 0.59     & 1747.59   \\
Naval    & $>10^6$     & $>10^6$      & 0.94  & 2.59  & 0.08  & 0.13  & $>10^6$    & 0.0  \\
Yacht    & 5.81   & 5326    & 0.21  & 0.83  & 0.69  & 0.30  & 0.30    & $>10^6$   \\
Kin8nm   & 33.9   & 0.15    & 0.26  & 0.14  & 0.08  & 0.34  & 0.02   & 0.1   \\
\bottomrule
\end{tabular}
\label{tab:std_dev_2}
\end{table*}

\begin{table*}[t]
\captionsetup{justification=centering}
\caption{Standard deviations of the average RMSE relative error on non-normalized training data}
\centering
\Large
\setlength{\tabcolsep}{5pt}
\begin{tabular}{lcccccccc}
\toprule
Dataset  & PBP    & Dropout & SVI   & KBNN  & TAGI  & HABNN & MCMC  & Laplace       \\
\midrule
Concrete   & 0.19      & 0.93  & 0.18    & 71.00 & 1.57   & 0.15  & 23.84 & 0.00         \\
Energy     & 0.26      & 0.36  & 0.15    & 67.60 & 3.14   & 0.67  & 34.94 & 0.01          \\
Wine       & 0.20      & 3.35  & 0.41    & 5.54  & 5.71   & 0.50  & 4.58  & 0.13          \\
Naval      & 0.01      & 0.30  & 0.05    & 318.55& 1.00   & 1.97  & 2.27  & 0.33          \\
Yacht      & 0.31      & 0.11  & 0.60    & 0.28  & 0.10   & 0.04  & 2.89  & 0.03          \\
Kin8nm     & 0.01      & 1.56  & 2.76    & 2.31  & 1.03   & 0.56  & 5.93  & 1.13          \\
\bottomrule
\end{tabular}
\label{tab:std_dev_rmse_rel_error}
\end{table*}

\begin{table*}[t]
\captionsetup{justification=centering}
\caption{Standard deviations of the average NLL relative error on non-normalized training data}
\centering
\Large
\setlength{\tabcolsep}{4pt}
\begin{tabular}{lcccccccc}
\toprule
Dataset   & PBP   & Dropout & SVI   & KBNN  & TAGI    & HABNN & MCMC  & Laplace \\
\midrule
Concrete  & 0.03  & 2.38    & 0.00  & 0.06  & 7.03    & 0.13  & 1.61  & 0.69    \\
Energy    & 0.11  & 0.88    & 0.02  & 0.00  & 5.82    & 0.40  & 11.58 & 0.68    \\
Wine      & 0.13  & 19.92   & 0.05  & 0.06  & 48.29   & 0.29  & 5.45  & 0.50    \\
Naval     & $>10^6$    & 0.33    & 0.06  & 0.10  & 686.40  & 0.65  & 0.31  & 1.96    \\
Yacht     & 0.14  & 0.24    & 0.18  & 0.20  & 203.05  & 0.27  & 0.46  & 0.95    \\
Kin8nm    & 2.60  & 118.21  & 3.88  & 8.21  & 1.28    & 0.22  & 7.67  & 5.52    \\
\bottomrule
\end{tabular}
\label{tab:std_dev_nll_rel_error}
\end{table*}

\section{Additional Explanation to the Industrial Benchmark}
\label{sec: add_exp_indstr_bench_}

In our Industrial Benchmark (IB) experiments, we define the per‐timestep loss as
\[
\mathcal{L}_t = -\,c_t \;-\; 3\,f_t,
\]
where \(c_t\) denotes consumption and \(f_t\) fatigue.  
With the environment set to the worst‐case setpoint of 100, an effective loss typically falls in the range 230–400.

For comparison, the original IB authors report achieving a loss of approximately 230 after training for over 100,000 timesteps on 96 CPUs—using histories of 10–30 past observations and roughly one week of runtime \cite{SIB_optimized}.  
By contrast, our HABNN model uses only the current sample (no history), a single hidden layer of 80 units, degrees of freedom initialized at 12, and a weight‐scale of 0.01. 
Despite this much lighter training regime, we are able to approach competitive performance.  
The results presented have been averaged over 100 runs.

\end{document}